\documentclass[11pt,a4paper]{article}
\usepackage[hyperref]{acl2020/acl2020}
\usepackage{times}
\usepackage{latexsym}

\usepackage{microtype}

\aclfinalcopy 
\def\aclpaperid{418} 

\usepackage{url}
\usepackage{hyperref}       
\usepackage{booktabs}       
\usepackage{amsfonts}       
\usepackage{nicefrac}       
\usepackage{microtype}   
\usepackage{boldline}
\usepackage{amsmath}
\usepackage{amssymb}
\usepackage{comment}
\usepackage{tabularx}
\usepackage[export]{adjustbox}
\usepackage{mathtools}
\usepackage{graphicx}
\usepackage{multirow}
\usepackage{subcaption}
\usepackage{tikz}
\usetikzlibrary{arrows}
\usetikzlibrary{positioning}
\usepackage{enumitem}
\usepackage{ctable}
\newcommand{\cmark}{\ding{51}}%
\newcommand{\xmark}{\ding{55}}%
\usepackage{wrapfig}
\usepackage{pifont}
\usepackage{color}
\usepackage{floatrow}
\DeclareFloatFont{footnotesize}{\footnotesize}
\newcommand{\kg}{\textsc{KG}{}}

\newcommand{\model}[1]{\textsc{AttH}}

\usepackage[symbol]{footmisc}

\title{
Low-Dimensional Hyperbolic Knowledge Graph Embeddings}

\author{Ines Chami$^1$\thanks{\ \ Work partially done during an internship at Google.}{\normalfont$\ $,} Adva Wolf$^1${\normalfont,} Da-Cheng Juan$^2${\normalfont,} Frederic Sala$^1${\normalfont,} Sujith Ravi$^3$\thanks{\ \ Work done while at Google AI.}$\ \ ${\normalfont and} Christopher R\'e{\normalfont$^1$} \\
  $^1$Stanford University \\
  $^2$Google Research \\
  $^3$Amazon Alexa \\
  \texttt{\{chami,advaw,fredsala,chrismre\}@cs.stanford.edu}\\
  \texttt{dacheng@google.com}\\
  \texttt{sravi@sravi.org}
  }

\date{}

\begin{document}
\maketitle
\begin{abstract}
  Knowledge graph (KG) embeddings learn low-dimensional representations of entities and relations to predict missing facts. 
KGs often exhibit hierarchical and logical patterns which must be preserved in the embedding space. 
For hierarchical data, hyperbolic embedding methods have shown promise for high-fidelity and parsimonious representations. 
However, existing hyperbolic embedding methods do not account for the rich logical patterns in KGs. 
In this work, we introduce a class of hyperbolic KG embedding models that simultaneously capture hierarchical and logical patterns. 
Our approach combines hyperbolic reflections and rotations with attention to model complex relational patterns.   
Experimental results on standard KG benchmarks show that our method improves over previous Euclidean- and hyperbolic-based efforts by up to 6.1\% in mean reciprocal rank (MRR) in low dimensions. 
Furthermore, we observe that different geometric transformations capture different types of relations while attention-based transformations generalize to multiple relations. 
In high dimensions, our approach yields new state-of-the-art MRRs of 49.6\% on WN18RR and 57.7\% on YAGO3-10. 
\end{abstract}

\section{Introduction}
Knowledge graphs (KGs), consisting of (\textit{head entity}, \textit{relationship}, \textit{tail entity}) triples, are popular data structures for representing factual knowledge to be queried and used in downstream applications such as word sense disambiguation, question answering, and information extraction. 
Real-world KGs such as Yago \cite{suchanek2007yago} or Wordnet \cite{miller1995wordnet} are usually incomplete, so a common approach to predicting missing links in KGs is via embedding into vector spaces.
Embedding methods learn representations of entities and relationships that preserve the information found in the graph, and have achieved promising results for many tasks. 

\begin{figure}[t]
\centering
\begin{subfigure}[b]{0.85\textwidth}
        \includegraphics[width=\textwidth]{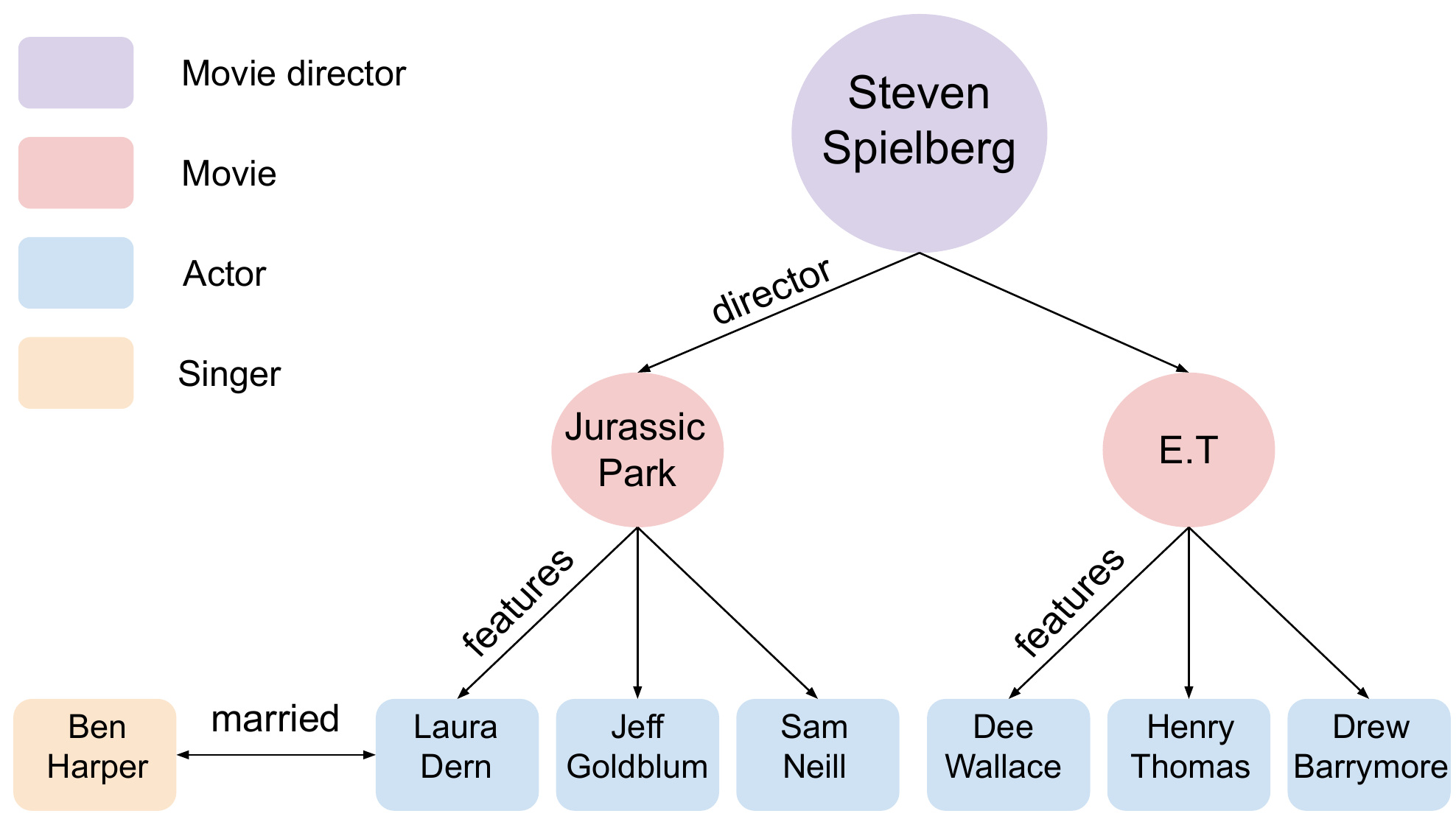}
    \end{subfigure}
\caption{
A toy example showing how KGs can simultaneously exhibit hierarchies and logical patterns. 
}\label{fig:toy_examples}
\end{figure}

Relations found in KGs have differing properties: for example, (\textit{Michelle Obama}, \textit{married to}, \textit{Barack Obama}) is symmetric, whereas hypernym relations like (\textit{cat}, \textit{specific type of}, \textit{feline}), are not (Figure \ref{fig:toy_examples}). 
These distinctions present a challenge to embedding methods: preserving each type of behavior requires producing a different geometric pattern in the embedding space. 
One popular approach is to use extremely high-dimensional embeddings, which offer more flexibility for such patterns. 
However, given the large number of entities found in KGs, doing so yields very high memory costs. 

For hierarchical data, hyperbolic geometry offers an exciting approach to learn low-dimensional embeddings while preserving latent hierarchies. 
Hyperbolic space can embed trees with arbitrarily low distortion in just two dimensions.
Recent research has proposed embedding hierarchical graphs into these spaces instead of conventional Euclidean space \cite{Nickel2017-mw,sala2018representation}. 
However, these works focus on embedding simpler graphs (e.g., weighted trees) and cannot express the diverse and complex relationships in \kg{}s.

We propose a new hyperbolic embedding approach that captures such patterns to achieve the best of both worlds. 
Our proposed approach produces the parsimonious representations offered by hyperbolic space, especially suitable for hierarchical relations, and is effective even with low-dimensional embeddings. 
It also uses rich transformations to encode logical patterns in KGs, previously only defined in Euclidean space. 
To accomplish this, we
(1) train hyperbolic embeddings with relation-specific curvatures to preserve multiple hierarchies in KGs; (2) parameterize hyperbolic isometries (distance-preserving operations) and leverage their geometric properties to capture relations' logical patterns, such as symmetry or anti-symmetry;
(3) and use a notion of hyperbolic attention to combine geometric operators and capture multiple logical patterns.

We evaluate the performance of our approach, \textsc{AttH}, on the KG link prediction task using the standard WN18RR \cite{dettmers2018convolutional,bordes2013translating}, FB15k-237 \cite{toutanova2015observed} and YAGO3-10 \cite{mahdisoltani2013yago3} benchmarks. 
(1) In low (32) dimensions, we improve over Euclidean-based models by up to 6.1\% in the mean reciprocical rank (MRR) metric. 
    In particular, we find that hierarchical relationships, such as WordNet's \textit{hypernym} and \textit{member meronym}, significantly benefit from hyperbolic space; we observe a 16\% to 24\% relative improvement versus Euclidean baselines.
    (2) We find that geometric properties of hyperbolic isometries directly map to logical properties of relationships.
    We study symmetric and anti-symmetric patterns and find that reflections capture symmetric relations while rotations capture anti-symmetry.
    (3) We show that attention based-transformations have the ability to generalize to multiple logical patterns. 
    For instance, we observe that \model{} recovers reflections for symmetric relations and rotations for the anti-symmetric ones.

In high (500) dimensions, we find that both hyperbolic and Euclidean embeddings achieve similar performance, and our approach achieves new state-of-the-art results (SotA), obtaining 49.6\% MRR on WN18RR and 57.7\% YAGO3-10. 
Our experiments show that trainable curvature is critical to generalize hyperbolic embedding methods to high-dimensions.
Finally, we visualize embeddings learned in hyperbolic spaces and show that hyperbolic geometry effectively preserves hierarchies in KGs. 

\section{Related Work}
Previous methods for KG embeddings also rely on geometric properties. 
Improvements have been obtained by exploiting either more sophisticated spaces (e.g., going from Euclidean to complex or hyperbolic space) or more sophisticated operations (e.g., from translations to isometries, or to learning graph neural networks). In contrast, our approach takes a step forward in both directions.

\paragraph{Euclidean embeddings} 
In the past decade, there has been a rich literature on Euclidean embeddings for \kg{} representation learning. 
These include translation approaches \cite{bordes2013translating,ji2015knowledge,wang2014knowledge,lin2015learning} or tensor factorization methods such as RESCAL \cite{nickel2011three} or DistMult \cite{yang2014embedding}. 
While these methods are fairly simple and have few parameters, they fail to encode important logical properties (e.g., translations can't encode symmetry).

\paragraph{Complex embeddings} 
Recently, there has been interest in learning embeddings in complex space, as in the {ComplEx} \cite{trouillon2016complex} and {RotatE} \cite{sun2019rotate} models. 
{RotatE} learns rotations in complex space, which are very effective in capturing logical properties such as symmetry, anti-symmetry, composition or inversion. 
The recent QuatE model \cite{zhang2019quaternion} learns KG embeddings using quaternions. 
However, a downside is that these embeddings require very high-dimensional spaces, leading to high memory costs. 

\paragraph{Deep neural networks} 
Another family of methods uses neural networks to produce KG embeddings. 
For instance, R-GCN \cite{schlichtkrull2018modeling} extends graph neural networks to the multi-relational setting by adding a relation-specific aggregation step. 
ConvE and ConvKB \cite{dettmers2018convolutional,Nguyen2018} leverage the expressiveness of convolutional neural networks to learn entity embeddings and relation embeddings. 
More recently, the KBGAT \cite{KBGAT2019} and A2N \cite{bansal2019a2n} models use graph attention networks for knowledge graph embeddings. 
A downside of these methods is that they are computationally expensive as they usually require pre-trained KG embeddings as input for the neural network. 

\paragraph{Hyperbolic embeddings}
To the best of our knowledge, {MuRP} \cite{balavzevic2019multi} is the only method that learns KG embeddings in hyperbolic space in order to target hierarchical data. 
{MuRP} minimizes hyperbolic distances between a re-scaled version of the head entity embedding and a translation of the tail entity embedding. 
It achieves promising results using hyperbolic embeddings with fewer dimensions than its Euclidean analogues. 
However, {MuRP} is a translation model and fails to encode some logical properties of relationships.
Furthermore, embeddings are learned in a hyperbolic space with fixed curvature, potentially leading to insufficient precision, and training relies on cumbersome Riemannian optimization. 
Instead, our proposed method leverages expressive hyperbolic isometries to simultaneously capture logical patterns and hierarchies. 
Furthermore, embeddings are learned using tangent space (i.e., Euclidean) optimization methods and trainable hyperbolic curvatures per relationship, avoiding precision errors that might arise when using a fixed curvature, and providing flexibility to encode multiple hierarchies. 

\section{Problem Formulation and Background}
We describe the KG embedding problem setting and give some necessary background on hyperbolic geometry. 
\subsection{Knowledge graph embeddings}
In the KG embedding problem, we are given a set of triples $(h, r, t)\in\mathcal{E} \subseteq \mathcal{V}\times \mathcal{R}\times\mathcal{V}$, where $\mathcal{V}$ and $\mathcal{R}$ are entity and relationship sets,  respectively.
The goal is to map entities $v\in\mathcal{V}$ to embeddings $\mathbf{e}_v\in\mathcal{U}^{d_\mathcal{V}}$ and relationships $r\in\mathcal{R}$ to embeddings $\mathbf{r}_r\in\mathcal{U}^{d_\mathcal{R}}$, for some choice of space $\mathcal{U}$ (traditionally  $\mathbb{R}$),
such that the \kg{} structure is preserved. 

Concretely, the data is split into $\mathcal{E}_{Train}$ and $\mathcal{E}_{Test}$ triples. 
Embeddings are learned by optimizing a scoring function $s:\mathcal{V}\times \mathcal{R}\times\mathcal{V}\rightarrow\mathbb{R}$, which measures triples' likelihoods.
$s(\cdot, \cdot, \cdot)$ is trained using triples in $\mathcal{E}_{Train}$ and the learned embeddings are then used to predict scores for triples in $\mathcal{E}_{Test}$. 
The goal is to learn embeddings such that the scores of triples in $\mathcal{E}_{Test}$ are high compared to triples that are not present in $\mathcal{E}$. 

\begin{figure}[t]
\centering
\begin{tikzpicture}[scale=0.4, transform shape]
  \pgfdeclareimage{img}{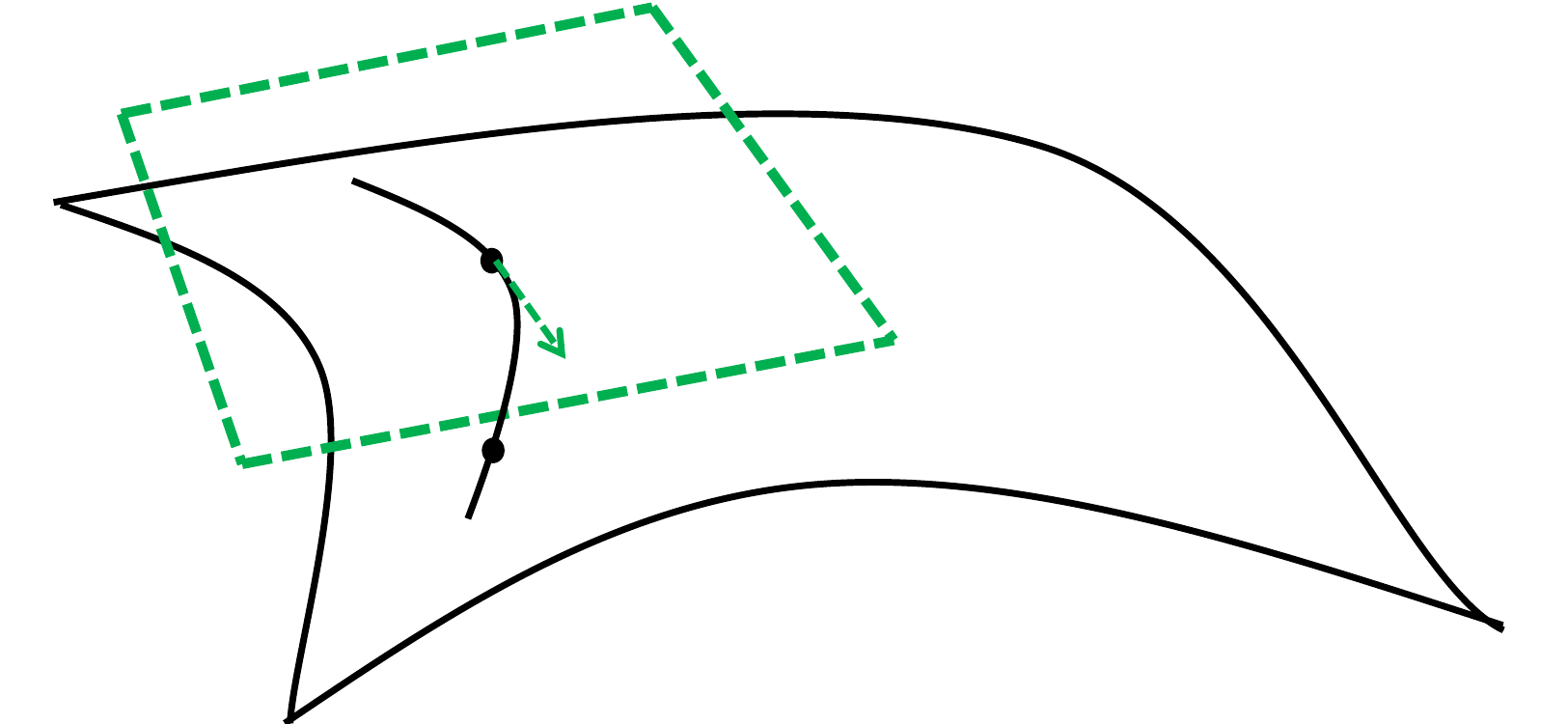}
  \node (img1) at (0,0) {\pgfuseimage{img}};
  \node (tspname) at (-2.0, 2.9) {\huge $\mathcal{T}_\mathbf{x}M$};
  \node (manname) at (6,-1.8)  {\huge $M$};
  \node (expname) at (-1.7, -1.1) {\huge  $\mathrm{exp}_\mathbf{x}(\mathbf{v})$};
  \node (vecname) at (-2.0, 0.1) {\huge  $\mathbf{v}$};
  \node (xname) at (-2.5, 1.1) {\huge  $\mathbf{x}$};
\end{tikzpicture}

\caption{An illustration of the exponential map  $\mathrm{exp}_\mathbf{x}(\mathbf{v})$, which maps the tangent space  $\mathcal{T}_\mathbf{x}M$ at the point $\mathbf{x}$ to the hyperbolic manifold $M$.
}
\label{fig:expmap}
\end{figure}
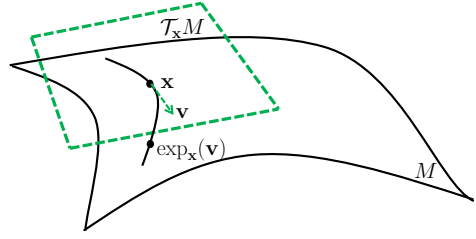 

\subsection{Hyperbolic geometry}\label{subsec:hyp_geom}
We briefly review key notions from hyperbolic geometry; a more in-depth treatment is available in standard texts \cite{robbin2011introduction}. 
Hyperbolic geometry is a non-Euclidean geometry with constant negative curvature. 
In this work, we use the $d$-dimensional Poincar\'e ball model with negative curvature $-c$ ($c>0$): $\mathbb{B}^{d,c}=\{\mathbf{x}\in\mathbb{R}^d:||\mathbf{x}||^2<\frac{1}{c}\}$, where $||\cdot||$ denotes the $L_2$ norm. 
For each point $\mathbf{x}\in\mathbb{B}^{d,c}$, the tangent space $\mathcal{T}^c_\mathbf{x}$ is a $d$-dimensional vector space containing all possible directions of paths in $\mathbb{B}^{d,c}$ leaving from $\mathbf{x}$.

The tangent space $\mathcal{T}^c_\mathbf{x}$ maps to $\mathbb{B}^{d,c}$ via the exponential map  (Figure \ref{fig:expmap}), and conversely, the logarithmic map maps $\mathbb{B}^{d,c}$ to $\mathcal{T}^c_\mathbf{x}$. 
In particular, we have closed-form expressions for these maps at the origin:
\begin{align}
\mathrm{exp}_\mathbf{0}^c(\mathbf{v})&=\mathrm{tanh}(\sqrt{c}||\mathbf{v}||)\frac{\mathbf{v}}{\sqrt{c}||\mathbf{v}||},\label{eq:expmap}\\
\mathrm{log}_\mathbf{0}^c(\mathbf{y})&=\mathrm{arctanh}(\sqrt{c}||\mathbf{y}||)\frac{\mathbf{y}}{\sqrt{c}||\mathbf{y}||}.\label{eq:logmap}
\end{align}
Vector addition is not well-defined in the hyperbolic space (adding two points in the Poincar\'e ball might result in a point outside the ball). 
Instead, M\"obius addition $\oplus^c$ \cite{ganea2018hyperbolicNN} provides an analogue to Euclidean addition for hyperbolic space. 
We give its closed-form expression in Appendix \ref{sec:appendix_hyp}. 
Finally, the hyperbolic distance on $\mathbb{B}^{d,c}$ has the explicit formula:
\begin{align}
d^c(\mathbf{x},\mathbf{y})=\frac{2}{\sqrt{c}}\mathrm{arctanh}(\sqrt{c}||-\mathbf{x}\oplus^c\mathbf{y}||).\label{eq:hyp_distance}
\end{align}

\section{Methodology} 
The goal of this work is to learn parsimonious hyperbolic embeddings that can encode complex logical patterns such as symmetry, anti-symmetry, or inversion while preserving latent hierarchies. 
Our model, \model{}, (1) learns KG embeddings in hyperbolic space in order to preserve hierarchies (Section \ref{subsec:hypmodel}), (2) uses a class of hyperbolic isometries parameterized by compositions of Givens transformations to encode logical patterns (Section \ref{subsec:isom}), (3) combines these isometries with hyperbolic attention (Section \ref{subsec:attention}). 
We describe the full model in Section \ref{subsec:fullmodel}.

\subsection{Hierarchies in hyperbolic space} \label{subsec:hypmodel}
As described, hyperbolic embeddings enable us to represent hierarchies even when we limit ourselves to low-dimensional spaces. 
In fact, two-dimensional hyperbolic space can represent any tree with arbitrarily small error \cite{sala2018representation}.

It is important to set the curvature of the hyperbolic space correctly. 
This parameter provides flexibility to the model, as it determines whether to embed relations into a more curved hyperbolic space (more ``tree-like"), or into a flatter, more ``Euclidean-like" geometry. 
For each relation, we learn a relation-specific absolute curvature $c_r$, enabling us to represent a variety of hierarchies. 
As we show in Section \ref{subsec:high_dim}, fixing, rather than learning curvatures can lead to significant performance degradation.

\subsection{Hyperbolic isometries}\label{subsec:isom}
Relationships often satisfy particular properties, such as symmetry: e.g., if (\textit{Michelle Obama}, \textit{married to}, \textit{Barack Obama}) holds, then (\textit{Barack Obama}, \textit{married to}, \textit{Michelle Obama}) does as well. 
These rules are not universal. For instance, (\textit{Barack Obama}, \textit{born in}, \textit{Hawaii}) is not symmetric.

Creating and curating a set of deterministic rules is infeasible for large-scale KGs; instead, embedding methods represent relations as parameterized geometric operations that directly map to logical properties. We use two such operations in hyperbolic space: \emph{rotations}, which effectively capture compositions or anti-symmetric patterns, and \emph{reflections}, which naturally encode symmetric patterns.

\paragraph{Rotations} Rotations have been successfully used to encode compositions in complex space with the RotatE model \cite{sun2019rotate}; we lift these to hyperbolic space. 
Compared to translations or tensor factorization approaches which can only infer some logical patterns, rotations can simultaneously model and infer \textit{inversion}, \textit{composition}, \textit{symmetric} or \textit{anti-symmetric} patterns.

\paragraph{Reflections} 
These isometries reflect along a fixed subspace.
While some rotations can represent symmetric relations (more specifically $\pi-$rotations), any reflection can naturally represent
symmetric relations, since their second power is the
identity. 
They provide a way to fill-in missing entries in symmetric triples, by applying the same operation to both the tail and the head entity.
For instance, by modelling \textit{sibling of} with a reflection, we can directly infer (\textit{Bob}, \textit{sibling of}, \textit{Alice}) from (\textit{Alice}, \textit{sibling of}, \textit{Bob}) and vice versa.

\begin{figure}
\centering
\begin{subfigure}[b]{0.49\textwidth}
        \includegraphics[width=\textwidth]{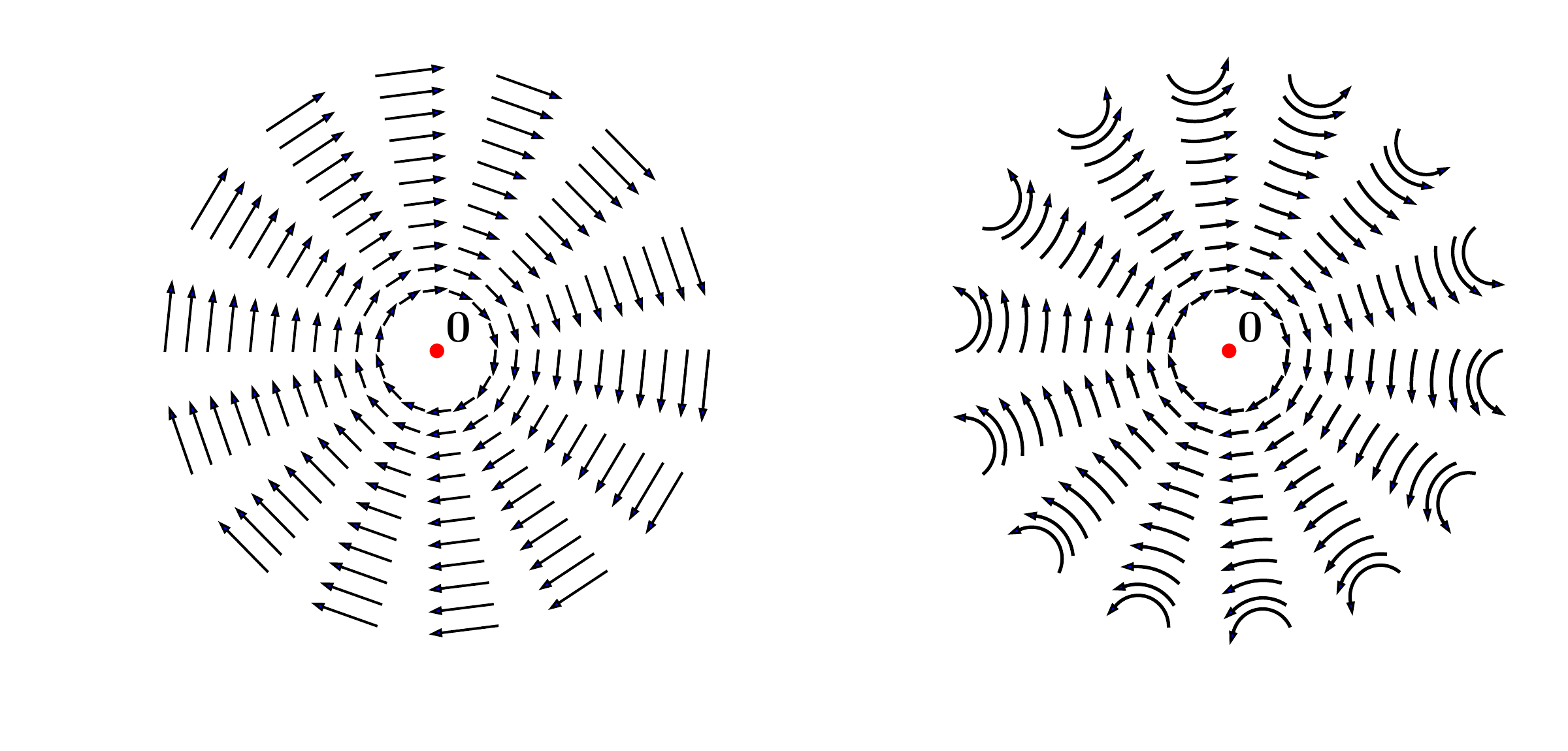}
        \caption{Rotations}
        \label{fig:rotations}
    \end{subfigure}
    \begin{subfigure}[b]{0.49\textwidth}
        \includegraphics[width=\textwidth]{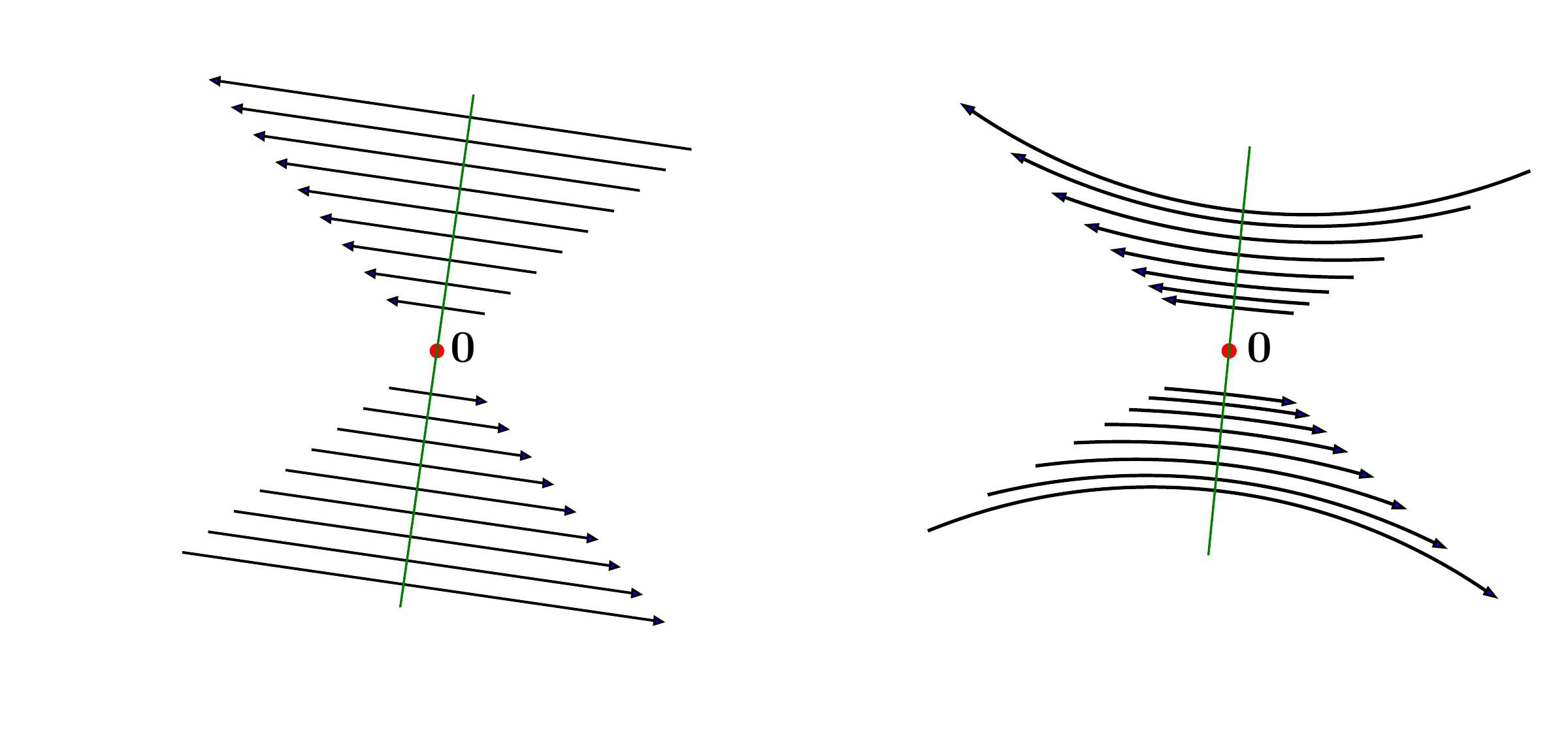}
        \caption{Reflections}
        \label{fig:reflections}
    \end{subfigure}
\caption{Euclidean (left) and hyperbolic (right) isometries. 
In hyperbolic space, the distance between start and end points after applying rotations or reflections is much larger than the Euclidean distance; it approaches the sum of the distances between the points and the origin, giving more ``room" to separate embeddings. This is similar to trees, where the shortest path between two points goes through their nearest common ancestor. 
}\label{fig:isometries}
\end{figure}

\paragraph{Parameterization} Unlike RotatE which models rotations via unitary complex numbers, we learn relationship-specific isometries using Givens transformations, $2\times2$ matrices commonly used in numerical linear algebra. 
Let $\Theta_r\coloneqq(\theta_{r,i})_{i\in\{1,\ldots\frac{d}{2}\}}$ and $\Phi_r\coloneqq(\phi_{r,i})_{i\in\{1,\ldots\frac{d}{2}\}}$ denote relation-specific parameters. 
Using an even number of dimensions $d$, our model parameterizes rotations and reflections with block-diagonal matrices of the form:
\begin{align}
    \mathrm{Rot}(\Theta_r)=\mathrm{diag}(G^+(\theta_{r,1}), \ldots, G^+(\theta_{r,\frac{d}{2}})),\label{eq:rotation} \\
    \mathrm{Ref}(\Phi_r)=\mathrm{diag}(G^-(\phi_{r,1}), \ldots, G^-(\phi_{r,\frac{n}{2}})), \label{eq:reflection} \\
    \text{where}\ \ G^{\pm}(\theta)\coloneqq \begin{bmatrix}
\mathrm{cos}(\theta) & \mp\mathrm{sin}(\theta) \\
\mathrm{sin}(\theta) & \ \ \pm\mathrm{cos}(\theta) \\
\end{bmatrix}.
\end{align}
Rotations and reflections of this form are hyperbolic isometries (distance-preserving).  
We can therefore directly apply them to hyperbolic embeddings while preserving the underlying geometry.  
Additionally, these transformations are computationally efficient and can be computed in linear time in the dimension.
We illustrate two-dimensional isometries in both Euclidean and hyperbolic spaces in Figure \ref{fig:isometries}. 

\subsection{Hyperbolic attention}\label{subsec:attention}
Of our two classes of hyperbolic isometries, one or the other may better represent a particular relation. 
To handle this, we use an attention mechanism to learn the right isometry. 
Thus we can represent symmetric, anti-symmetric or mixed-behaviour relations (i.e. neither symmetric nor anti-symmetric) as a combination of rotations and reflections.

Let $\mathbf{x}^H$ and $\mathbf{y}^H$ be hyperbolic points (e.g., reflection and rotation embeddings), and $\mathbf{a}$ be an attention vector. Our approach maps hyperbolic representations to tangent space representations, $\mathbf{x}^E=\mathrm{log}^c_\mathbf{0}(\mathbf{x}^H)$ and $\mathbf{y}^E=\mathrm{log}^c_\mathbf{0}(\mathbf{y}^H)$, and computes attention scores:
\begin{align*}
(\alpha_\mathbf{x}, \alpha_\mathbf{y})=\mathrm{Softmax}(\mathbf{a}^T\mathbf{x}^E, \mathbf{a}^T\mathbf{y}^E).
\end{align*}
We then compute a weighted average using the recently proposed tangent space average \cite{chami2019hyperbolic,liu2019hyperbolic}: 
\begin{align}
\mathrm{Att}(\mathbf{x}^H, \mathbf{y}^H;\mathbf{a})\coloneqq\mathrm{exp}_\mathbf{0}^c(\alpha_\mathbf{x}\mathbf{x}^E+\alpha_\mathbf{y}\mathbf{y}^E).\label{eq:attention}
\end{align}

\subsection{The \model{} model} \label{subsec:fullmodel}
We have all of the building blocks for \model{}, and can now describe the model architecture. 
Let $(\mathbf{e}_v^H)_{v\in\mathcal{V}}$ and $(\mathbf{r}_r^H)_{r\in\mathcal{R}}$ denote entity and relationship hyperbolic embeddings respectively. 
For a triple $(h, r, t)\in\mathcal{V}\times\mathcal{R}\times\mathcal{V}$, \model{} applies relation-specific rotations (Equation \ref{eq:rotation}) and reflections (Equation \ref{eq:reflection}) to the head embedding: 
\begin{align}
\mathbf{q}^H_{\mathrm{Rot}}=\mathrm{Rot}(\Theta_r)\mathbf{e}_h^H,\  \mathbf{q}^H_{\mathrm{ref}}=\mathrm{Ref}(\Phi_r)\mathbf{e}_h^H.
\end{align}
\model{} then combines the two representations using hyperbolic attention (Equation \ref{eq:attention}) and applies a hyperbolic translation:
\begin{equation}
\mathrm{Q}(h, r)=\mathrm{Att}(\mathbf{q}^H_{\mathrm{Rot}}, \mathbf{q}^H_{\mathrm{Ref}};\mathbf{a}_r)\oplus^{c_r}\mathbf{r}_r^H.
\end{equation}

Intuitively, rotations and reflections encode logical patterns while translations capture tree-like structures by moving between levels of the hierarchy. 
Finally, query embeddings are compared to target tail embeddings via the hyperbolic distance (Equation \ref{eq:hyp_distance}). 
The resulting scoring function is:
\begin{align}
    s(h, r, t)=-d^{c_r}(Q(h,r), \mathbf{e}_t^H)^2+b_h+b_t,\label{eq:score}
\end{align}
where $(b_v)_{v\in\mathcal{V}}$ are entity biases which act as margins in the scoring function \cite{tifrea2018poincar,balavzevic2019multi}. 

The model parameters are then $\{(\Theta_r, \Phi_r, \mathbf{r}_r^H, \mathbf{a}_r, c_r)_{r\in\mathcal{R}},(\mathbf{e}_v^H, b_v)_{v\in\mathcal{V}}\}$.
Note that the total number of parameters in \model{} is $\mathcal{O}(|\mathcal{V}|d)$, similar to traditional models that do not use attention or geometric operations.
The extra cost is proportional to the number of relations, which is usually much smaller than the number of entities. 

\begin{table}[t]
\centering
\begin{tabular}{lcccc}
\clineB{1-5}{2}
{Dataset} & {\#entities} & {\#relations} & {\#triples} & $\xi_G$ \ \\
\clineB{1-5}{2}
WN18RR & 41k & 11 & 93k & -2.54 \\
\hline
FB15k-237 & 15k & 237 & 310k & -0.65 \\
\hline
YAGO3-10 & 123k & 37 & 1M & -0.54 \\
\clineB{1-5}{2}
\end{tabular}
\caption{Datasets statistics. The lower the metric $\xi_G$ is, the more tree-like the knowledge graph is.} 
\label{tab:datasets}

\end{table}
\section{Experiments}\label{subsec:LP}
In low dimensions, we hypothesize (1) that hyperbolic embedding methods obtain better representations and allow for improved downstream performance for hierarchical data (Section \ref{sec:explowdim}).
(2) We expect the performance of relation-specific geometric operations to vary based on the relation's logical patterns (Section \ref{sec:expops}). 
(3) In cases where the relations are neither purely symmetric nor anti-symmetric, we anticipate that hyperbolic attention outperforms the models which are based on solely reflections or rotations  (Section \ref{sec:expatt}). 
Finally, in high dimensions, we expect hyperbolic models with trainable curvature to learn the best geometry, and perform similarly to their Euclidean analogues (Section \ref{subsec:high_dim}).
\subsection{Experimental setup}
\begin{table*}[t]
\resizebox{\textwidth}{!}{\renewcommand{\arraystretch}{1.1}
    \centering
    \begin{tabular}{@{}lllccccccccccc@{}}
    \clineB{1-14}{2}
    & & \multicolumn{4}{c}{WN18RR} & \multicolumn{4}{c}{FB15k-237} & \multicolumn{4}{c}{YAGO3-10} \\
    $\mathcal{U}$ & Model & MRR & H@1 & H@3 & H@10 & MRR & H@1 & H@3 & H@10 & MRR & H@1 & H@3 & H@10\\
     \clineB{1-14}{2}
     \multirow{2}{*}{{$\mathbb{R}^d$}} & RotatE & .387 & .330 & .417 & .491 & .290 & .208 & .316 & .458 & - & - &-&- \\
     & {MuRE} & .458 & \underline{{.421}} & .471 & .525 & .313 & .226 & .340 & .489 & .283 & .187 & .317 & .478 \\
     \cline{1-14}
     $\mathbb{C}^d$ & {ComplEx-N3} & .420 & .390 & .420 & .460 & .294 & .211 & .322 & .463 & \underline{.336} & \underline{.259} & \underline{.367} & \underline{.484} \\ 
     \cline{1-14}
     $\mathbb{B}^{d,1}$ & {MuRP} & \underline{{.465}} & .420 & \underline{.484} & \underline{{.544}} & \underline{{.323}} & \underline{{.235}} & \underline{{.353}} & \underline{\textbf{.501}} & .230 & .150 & .247 & .392 \\
     \clineB{1-14}{2}
    \multirow{3}{*}{{$\mathbb{R}^d$}} & \textsc{RefE} & .455 & .419 & .470 & .521 & .302 & .216 & .330 & .474 & .370 & .289 & .403 & .527 \\
    & \textsc{RotE} & .463 & .426 & .477 & .529 & .307 & .220 & .337 & .482 & {.381} & {.295} & {.417} & {.548} \\
     & \textsc{AttE} & .456 & .419 & .471 & .526 & .311 & .223 & .339 & .488 & {.374} & {.290} & {.410} & {.537}\\
     \cline{1-14}
     \multirow{3}{*}{{$\mathbb{B}^{d,c}$}}  & \textsc{RefH} & .447 & .408 & .464 & .518 & .312 & .224 & .342 & .489 & .381 & .302 & .415 & .530 \\ 
     & \textsc{RotH} & \textbf{.472} & \textbf{.428} & \textbf{.490} & \textbf{.553} & {.314} & {.223} & {.346} & {.497} & .393 & .307 & .435 & 559 \\
    & \textsc{AttH} & {.466} & .419 & {.484} & {.551} & \textbf{.324} & \textbf{.236} & \textbf{.354} & \textbf{.501} & \textbf{.397} & \textbf{.310} & \textbf{.437} & \textbf{.566}\\
    \clineB{1-14}{2}
    \end{tabular}
    }
    \caption{Link prediction results for low-dimensional embeddings ($d=32$) in the filtered setting. Best score in \textbf{bold} and best published \underline{underlined}. Hyperbolic isometries significantly outperform Euclidean baselines on WN18RR and YAGO3-10, both of which exhibit hierarchical structures.}
    \label{tab:low_dim}
\end{table*}
\paragraph{Datasets} 
We evaluate our approach on the link prediction task using three standard competition benchmarks, namely WN18RR \cite{bordes2013translating,dettmers2018convolutional}, FB15k-237 \cite{bordes2013translating,toutanova2015observed} and YAGO3-10 \cite{mahdisoltani2013yago3}. 
WN18RR is a subset of WordNet containing 11 lexical relationships between 40,943 word senses, and has a natural hierarchical structure, e.g., (\textit{car}, \textit{hypernym of}, \textit{sedan}).  
FB15k-237 is a subset of Freebase, a collaborative KB of general world knowledge. 
FB15k-237 has 14,541 entities and 237 relationships, some of which are non-hierarchical, such as \textit{born-in} or \textit{nationality}, while others have natural hierarchies, such as \textit{part-of} (for organizations). 
YAGO3-10 is a subset of YAGO3, containing 123,182 entities and 37
relations, where most relations provide descriptions of people. 
Some relationships have a hierarchical structure such as \textit{playsFor} or \textit{actedIn}, while others induce logical patterns, like \textit{isMarriedTo}. 

For each KG, we follow the standard data augmentation protocol by adding inverse relations \cite{lacroix2018canonical} to the datasets.
Additionally, we estimate the global graph curvature $\xi_G$ \cite{gu2019mixed} (see Appendix \ref{subsec:curvature_est} for more details), which is a distance-based measure of how close a given graph is to being a tree. 
We summarize the datasets' statistics in Table \ref{tab:datasets}.

\paragraph{Baselines} 
We compare our method to SotA models, including MurP \cite{balazevic2019tucker}, MurE (which is the Euclidean analogue or MurP), RotatE \cite{sun2019rotate}, ComplEx-N3 \cite{lacroix2018canonical} and TuckER \cite{balazevic2019tucker}. 
Baseline numbers in high dimensions (Table \ref{tab:high_dim}) are taken from the original papers, while baseline numbers in the low-dimensional setting (Table \ref{tab:low_dim}) are computed using open-source implementations of each model. 
In particular, we run hyper-parameter searches over the same parameters as the ones in the original papers to compute baseline numbers in the low-dimensional setting.

\begin{figure}[t]
    \centering
    \includegraphics[width=0.95\textwidth]{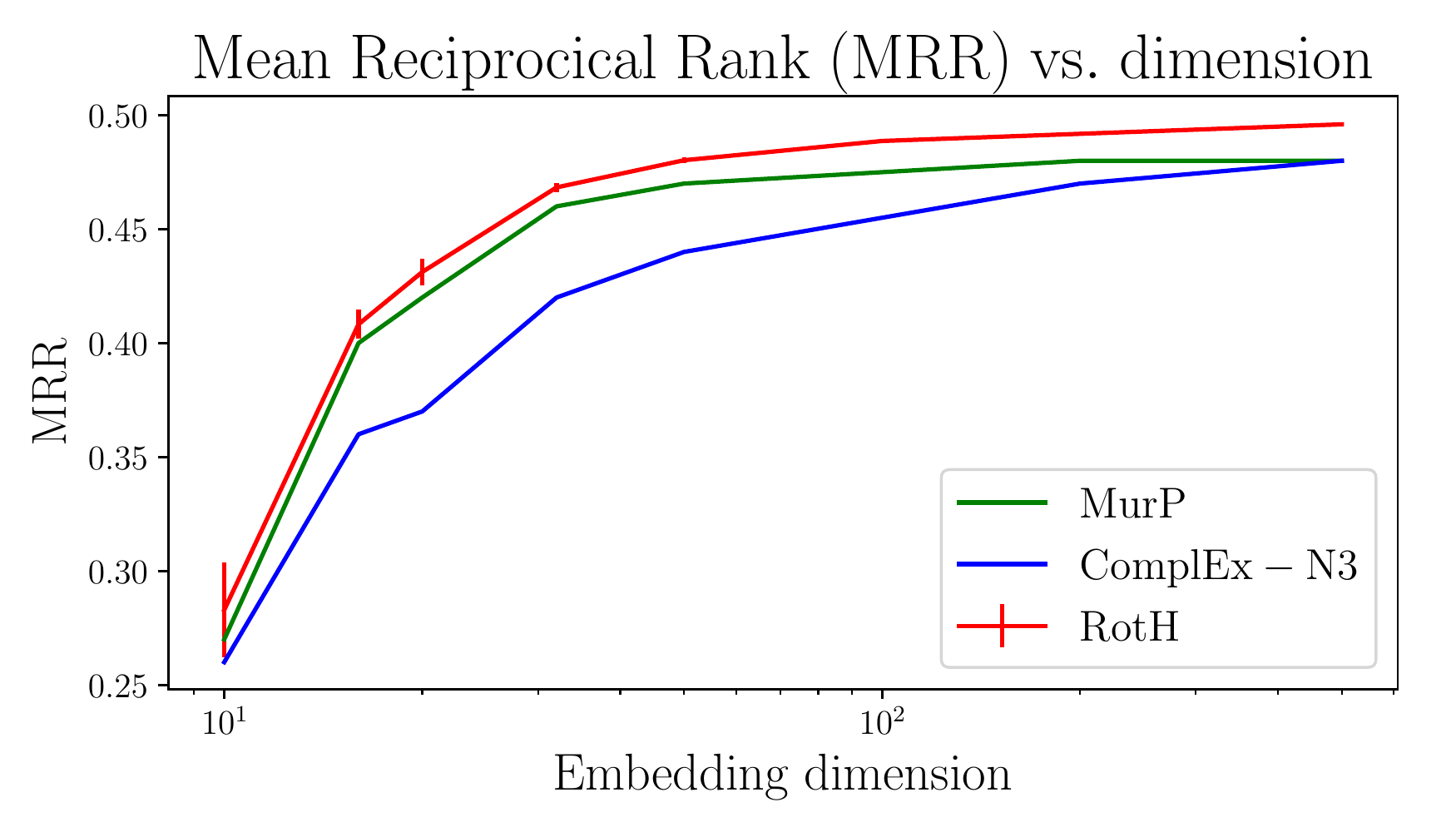}
    \caption{WN18RR MRR dimension for $d\in\{10, 16, 20, 32, 50, 200, 500\}$. Average and standard deviation computed over 10 runs for \textsc{RotH}.}
    \label{fig:low_dim_plot}
\end{figure}
\paragraph{Ablations} 
To analyze the benefits of hyperbolic geometry, we evaluate the performance of \textsc{AttE}, which is equivalent to \model{} with curvatures set to zero.
Additionally, to better understand the role of attention, we report scores for variants of \textsc{AttE}/\textsc{H} using only rotations (\textsc{RotE}/\textsc{H}) or reflections (\textsc{RefE}/\textsc{H}).

\paragraph{Evaluation metrics} 
At test time, we use the scoring function in Equation \ref{eq:score} to rank the correct tail or head entity against all possible entities, and use in use inverse relations for head prediction \cite{lacroix2018canonical}.
Similar to previous work, we compute two ranking-based metrics: (1) mean reciprocal rank (MRR), which measures the mean of inverse ranks assigned to correct entities, and (2) hits at $K$ (H@$K$, $K$ $\in\{1, 3, 10\}$), which measures the proportion of correct triples among the top $K$ predicted triples. 
We follow the standard evaluation protocol in the filtered setting \cite{bordes2013translating}: all true triples in the \kg{} are filtered out during evaluation, since predicting a low rank for these triples should not be penalized. 

\paragraph{Training procedure and implementation} 
We train \model{} by minimizing the full cross-entropy loss with uniform negative sampling, where negative examples for a triple $(h, r, t)$ are sampled uniformly from all possible triples obtained by perturbing the tail entity:   
\begin{align}
    \mathcal{L}=\sum\limits_{t'\sim\mathcal{U}(\mathcal{V})}\mathrm{log}(1 +& \mathrm{exp}(y_{t'}s(h,r,t'))),\label{eq:loss}\\
    \mathrm{where}\ \ y_{t'}&=\begin{cases}-1 & \mathrm{if}\ \  t'=t\\
    1 & \mathrm{otherwise.} 
    \end{cases}\nonumber
\end{align}
Since optimization in hyperbolic space is practically challenging, we instead define all parameters in the tangent space at the origin, optimize embeddings using standard Euclidean techniques, and use the exponential map to recover the hyperbolic parameters \cite{chami2019hyperbolic}. 
We provide more details on tangent space optimization in Appendix \ref{subsec:tangent_optim}. 
We conducted a grid search to select the learning rate, optimizer, negative sample size, and batch size, using the validation set to select the best hyperparameters.
Our best model hyperparameters are detailed in Appendix \ref{subsec:appendix_param}. 
We conducted all our experiments on NVIDIA Tesla P100 GPUs and make our implementation publicly available\footnote{Code available at \url{https://github.com/tensorflow/neural-structured-learning/tree/master/research/kg_hyp_emb}}.

\begin{table}[t]
\centering
\resizebox{\textwidth}{!}{\renewcommand{\arraystretch}{1.0}
\begin{tabular}{lccccc}
\clineB{1-6}{2}
{Relation} & {$\text{Khs}_G$} & {$\xi_G$} & {\textsc{RotE}} & {\textsc{RotH}} & {Improvement} \\
\clineB{1-6}{2}
 member meronym & 1.00& -2.90 & .320 & \textbf{.399} & 24.7\% \\
 hypernym & 1.00& -2.46 & .237 & \textbf{.276} & 16.5\% \\
 has part & 1.00& -1.43 & .291 & \textbf{.346} & 18.9\% \\
 instance hypernym & 1.00& -0.82 & .488 & \textbf{.520} & 6.56\% \\
 member of domain region & 1.00& -0.78 & \textbf{.385} & .365 & -5.19\% \\
 member of domain usage & 1.00& -0.74 & \textbf{.458} & .438 & -4.37\% \\
 synset domain topic of & 0.99& -0.69 & .425 & \textbf{.447} & 5.17\% \\
 also see & 0.36& -2.09 & .634 & \textbf{.705} & 11.2\% \\
 derivationally related form & 0.07& -3.84 & .960 & \textbf{.968} & 0.83\% \\
 similar to & 0.07& -1.00 & \textbf{1.00} & \textbf{1.00} & 0.00\% \\
 verb group & 0.07& -0.50 & \textbf{.974} & \textbf{.974} & 0.00\% \\
\clineB{1-6}{2}
\end{tabular}
\caption{Comparison of H@10 for WN18RR relations. Higher $\text{Khs}_G$ and lower $\xi_G$ means more hierarchical.} 
\label{tab:rel}
}
\end{table}
\subsection{Results in low dimensions} \label{sec:explowdim}
We first evaluate our approach in the low-dimensional setting for $d=32$, which is approximately one order of magnitude smaller than SotA Euclidean methods.
Table \ref{tab:low_dim} compares the performance of \model{} to that of other baselines, including the recent hyperbolic (but not rotation-based) {MuRP} model. 
In low dimensions, hyperbolic embeddings offer much better representations for hierarchical relations, confirming our hypothesis.
\model{} improves over previous Euclidean and hyperbolic methods by 0.7\% and 6.1\% points in MRR on WN18RR and YAGO3-10 respectively. 
Both datasets have multiple hierarchical relationships, suggesting that the hierarchical structure imposed by hyperbolic geometry leads to better embeddings. 
On FB15k-237, \model{} and MurP achieve similar performance, both improving over Euclidean baselines. 
We conjecture that translations are sufficient to model relational patterns in FB15k-237.

To understand the role of dimensionality, we also conduct experiments on WN18RR against SotA methods under varied low-dimensional settings (Figure \ref{fig:low_dim_plot}).
We include error bars for our method with average MRR and standard deviation computed over 10 runs. 
Our approach consistently outperforms all baselines, suggesting that hyperbolic embeddings still attain high-accuracy across a broad range of dimensions. 

\begin{figure*}[t]
\centering
\begin{subfigure}[b]{0.47\textwidth}
        \includegraphics[width=0.97\textwidth]{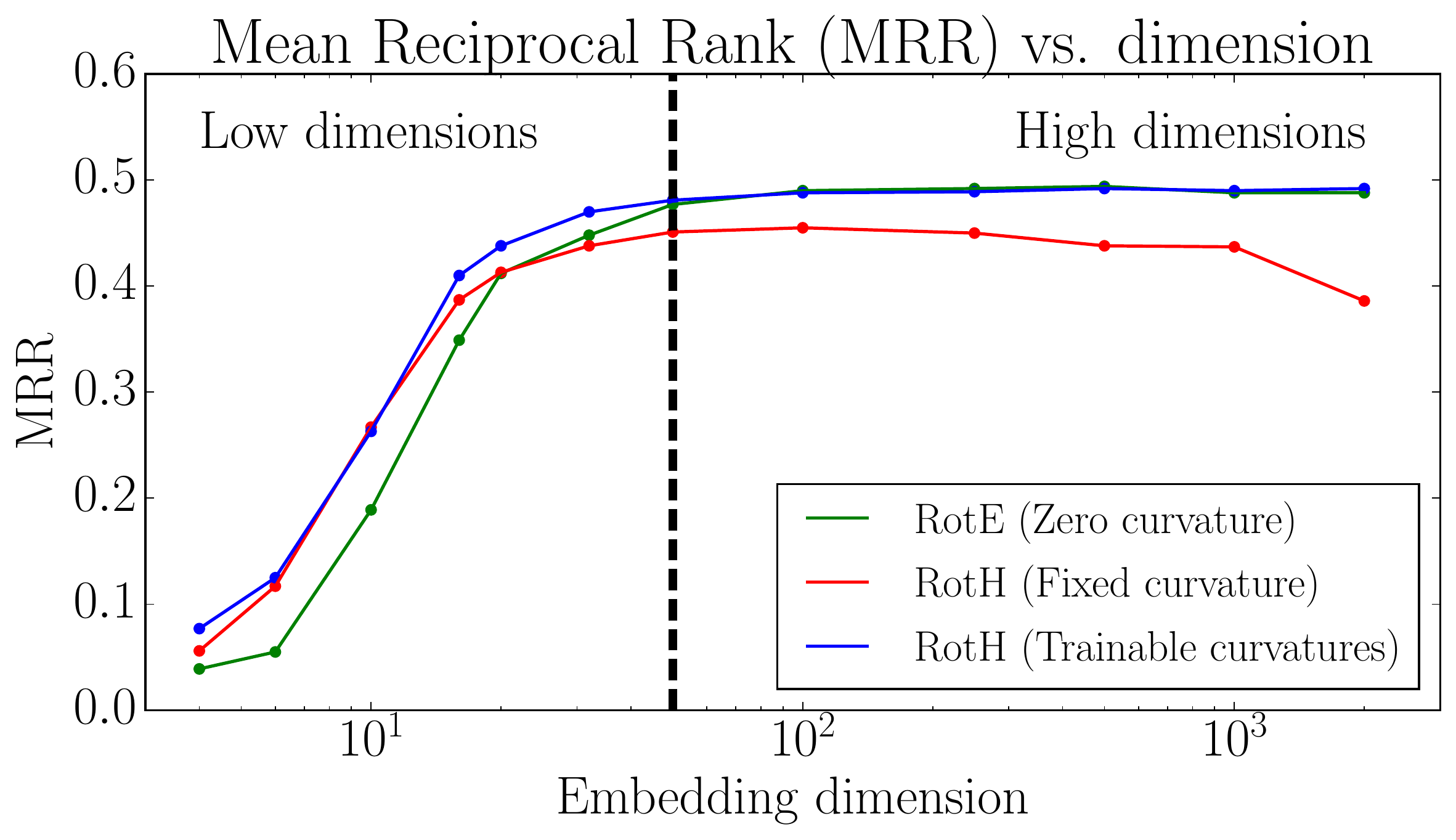}
        \caption{MRR for fixed and trainable curvatures on WN18RR.}
        \label{fig:sweep_d}
    \end{subfigure}
    \begin{subfigure}[b]{0.47\textwidth}
        \includegraphics[width=0.97\textwidth]{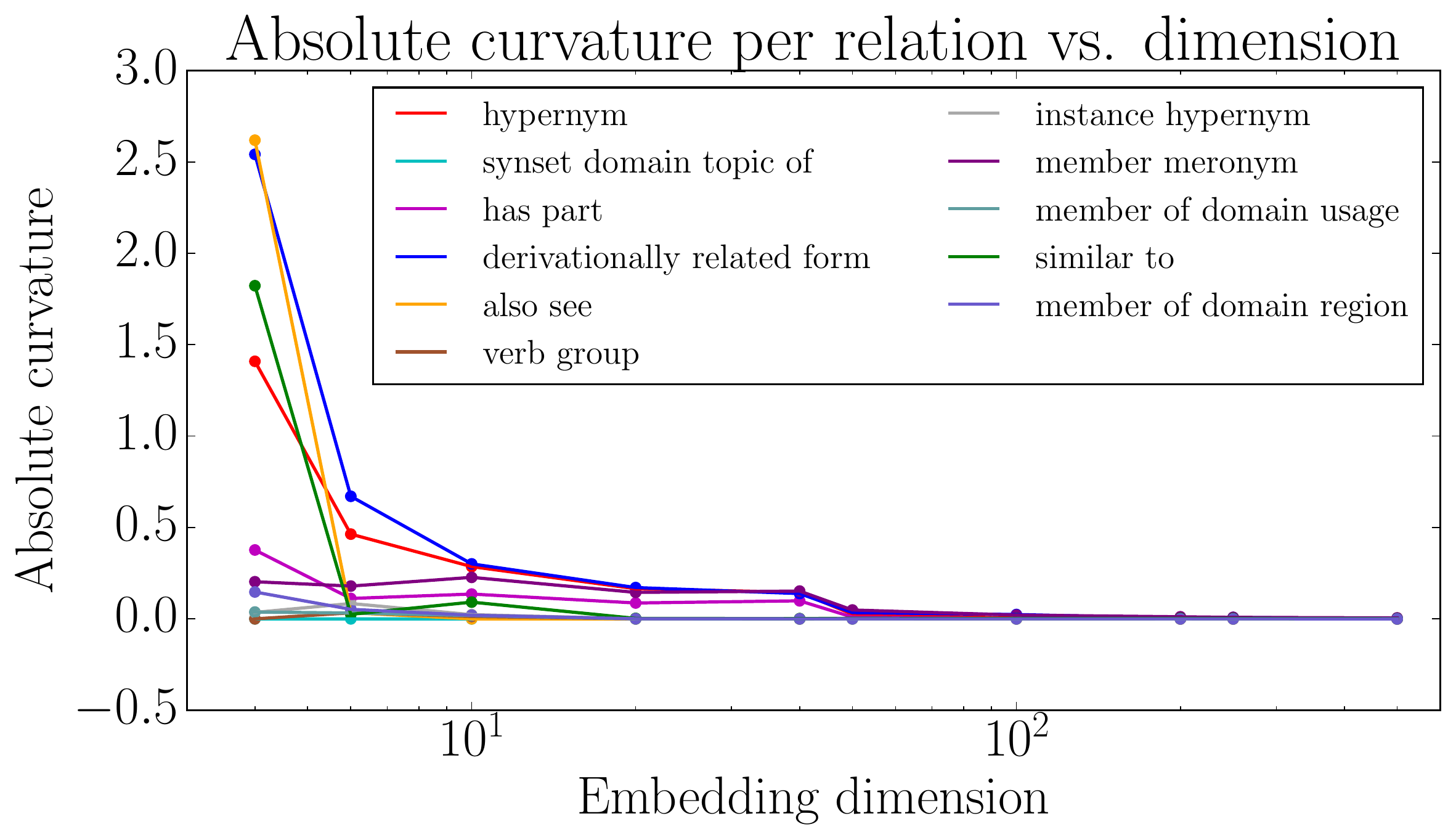}
        \caption{Curvatures learned by with \textsc{RotH} on WN18RR.}
        \label{fig:learned_c}
    \end{subfigure}
\caption{(a): \textsc{RotH} offers improved performance in low dimensions; in high dimensions, fixed curvature degrades performance, while trainable curvature approximately recovers Euclidean space. (b): As the dimension increases, the learned curvature of hierarchical relationships tends to zero.}\label{fig:curvature}
\end{figure*}

Additionally, we measure performance per relation on WN18RR in Table \ref{tab:rel} to understand the benefits of hyperbolic geometric on hierarchical relations. 
We report the Krackhardt hierarchy score (Khs$_G$) \cite{balavzevic2019multi} and estimated curvature per relation  (see Appendix \ref{subsec:curvature_est} for more details).
We consider a relation to be hierarchical when its corresponding graph is close to tree-like (low curvature, high Khs$_G$).
We observe that hyperbolic embeddings offer much better performance on hierarchical relations such as \textit{hypernym} or \textit{has part}, while Euclidean and hyperbolic embeddings have similar performance on non-hierarchical relations such as \textit{verb group}.
We also plot the learned curvature per relation versus the embedding dimension in Figure \ref{fig:learned_c}. 
We note that the learned curvature in low dimensions directly correlates with the estimated graph curvature $\xi_G$ in Table \ref{tab:rel}, suggesting that the model with learned curvatures learns more ``curved'' embedding spaces for tree-like relations. 

Finally, we observe that MurP achieves lower performance than MurE on YAGO3-10, while \model{} improves over \textsc{AttE} by 2.3\% in MRR.
This suggests that trainable curvature is critical to learn embeddings with the right amount of curvature, while fixed curvature might degrade performance. We elaborate further on this point in Section \ref{subsec:high_dim}.  

\begin{table}[t]
\centering
\resizebox{\textwidth}{!}{\renewcommand{\arraystretch}{1.}
\begin{tabular}{lccccc}
\clineB{1-6}{2}
{Relation} & {Anti-symmetric} & {Symmetric}  & {\textsc{RotH}} & {\textsc{RefH}} & {\textsc{AttH}} \\
\clineB{1-6}{2}
hasNeighbor & \xmark & \cmark & .750 & \textbf{1.00} & \textbf{1.00} \\
isMarriedTo & \xmark & \cmark & .941 & .941 & \textbf{1.00} \\
actedIn & \cmark & \xmark & .145 & .110 & \textbf{.150} \\
hasMusicalRole & \cmark & \xmark & .431 & .375 & \textbf{.458} \\
directed & \cmark & \xmark & .500 & .450 & \textbf{.567} \\
graduatedFrom & \cmark & \xmark & .262 & .167 & \textbf{.274} \\
playsFor & \cmark & \xmark & \textbf{.671} & .642 & .664 \\
wroteMusicFor & \cmark & \xmark & \textbf{.281} & .188 & .266 \\
hasCapital & \cmark & \xmark & .692 & \textbf{.731}& \textbf{.731} \\
dealsWith & \xmark & \xmark & .286 & .286 & \textbf{.429} \\
isLocatedIn	& \xmark & \xmark & .404 & .399 & \textbf{.420} \\
\clineB{1-6}{2}
\end{tabular} 
\caption{Comparison of geometric transformations on a subset of YAGO3-10 relations.} 
\label{tab:attention}
}
\end{table}

\subsection{Hyperbolic rotations and reflections} \label{sec:expops}
In our experiments, we find that rotations work well on WN18RR, which contains multiple hierarchical and anti-symmetric relations, while reflections work better for YAGO3-10 (Table \ref{tab:high_dim}). 
To better understand the mechanisms behind these observations, we analyze two specific patterns: relation symmetry and anti-symmetry.  
We report performance per-relation on a subset of YAGO3-10 relations in Table \ref{tab:attention}. 
We categorize relations into symmetric, anti-symmetric, or neither symmetric nor anti-symmetric categories using data statistics.
More concretely, we consider a relation to satisfy a logical pattern when the logical condition is satisfied by most of the triplets (e.g., a relation $r$ is symmetric if for most KG triples $(h, r, t)$, $(t, r, h)$ is also in the KG). 
We observe that reflections encode symmetric relations particularly well, while rotations are well suited for anti-symmetric relations.  
This confirms our intuition---and the motivation for our approach---that particular geometric properties capture different kinds of logical properties.

\subsection{Attention-based transformations} \label{sec:expatt}
One advantage of using relation-specific transformations is that each relation can learn the right geometric operators based on the logical properties it has to satisfy.  
In particular, we observe that in both low- and high-dimensional settings, attention-based models can recover the performance of the best transformation on all datasets (Tables \ref{tab:low_dim} and \ref{tab:high_dim}). 
Additionally, per-relationship results on YAGO3-10 in Table \ref{tab:attention} suggest that \model{} indeed recovers the best geometric operation. 

Furthermore, for relations that are neither symmetric nor anti-symmetric, we find that \model{} can outperform rotations and reflections, suggesting that combining multiple operators with attention can learn more expressive operators to model mixed logical patterns. 
In other words, attention-based transformations alleviate the need to conduct experiments with multiple geometric transformations by simply allowing the model to choose which one is best for a given relation. 

\begin{table*}[t]	
\resizebox{\textwidth}{!}{\renewcommand{\arraystretch}{1.1}
\centering
\begin{tabular}{@{}lllcccccccccccc@{}}
\clineB{1-14}{2} 
& & \multicolumn{4}{c}{WN18RR} & \multicolumn{4}{c}{FB15k-237} & \multicolumn{4}{c}{YAGO3-10}\\
$\mathcal{U}$ & Model & MRR & H@1 & H@3 & H@10 & MRR & H@1 & H@3 & H@10 & MRR & H@1 & H@3 & H@10 \\
\clineB{1-14}{2}
\multirow{4}{*}{{$\mathbb{R}^d$}} & {DistMult} & .430 & .390 & .440 & .490 & .241 & .155 & .263 & .419 & .340 & .240 & .380 & .540 \\  
& {ConvE} & .430 & .400 & .440 & .520 & .325 & .237 & .356 & .501 & .440 & .350 & .490 & .620 \\
& TuckER & .470 & .443 & .482 & .526 & \underline{{.358}} & \underline{{.266}} & \underline{{.394}} & .544 & - & - & - & - \\
& {MurE} & .475 & .436 & .487 & .554 & .336 & {.245} & .370 & .521 & .532 & .444 & .584 & .694 \\
\hline
\multirow{2}{*}{{$\mathbb{C}^d$}}
& {ComplEx-N3} & .480 & .435 & .495 & .572 & {{.357}} & {{.264}} & {{.392}} & {.547} & \underline{.569} & \underline{{.498}} & \underline{.609} & \underline{.701}\\
& {RotatE} & {.476} & .428 & {.492} & {.571} & {.338} & .241 & {.375} & {.533} & .495 & .402 & .550 & .670 \\  
\hline
$\mathbb{H}^d$ & {Quaternion} & \underline{.488} & \underline{.438} & \underline{.508} & \underline{.582} & {.348} & {.248} & {.382} & \underline{\textbf{.550}} & - & - & - & - \\ 
\hline
\multirow{1}{*}{{$\mathbb{B}^{d,1}$}} & {MurP} & {.481} & {.440} & {.495} & .566 & .335 & .243 & .367 & .518 & .354 & .249 & .400 & 567 \\ 
\clineB{1-14}{2}
\multirow{3}{*}{{$\mathbb{R}^d$}} & \textsc{RefE} & .473 & .430 & .485 & .561 & .351 & .256 & .390 & .541 & \textbf{.577} & \textbf{.503} & \textbf{.621} & \textbf{.712}\\
& \textsc{RotE} & {.494} & {.446} & {.512} & {.585} & .346 & .251 & .381  & .538 & .574 & .498 & \textbf{.621} & {.711}\\
& \textsc{AttE} & {.490} & {.443} & {.508} & .581 & .351 & .255 & .386 & .543 & {.575} & {.500} & \textbf{.621} & {.709} \\
\hline
\multirow{3}{*}{{$\mathbb{B}^{d,c}$}} &  \textsc{RefH} & .461 & .404 & .485 & .568 & .346 & .252 & .383 & .536 & .576 & .502 & .619 & .711\\
& \textsc{RotH} & \textbf{.496} & \textbf{.449} & \textbf{.514} & \textbf{.586} & .344 & .246 & .380 & .535 & .570 & .495 & .612 & .706 \\
 & \textsc{AttH} & .486 & .443 & .499 & .573 &  .348 & .252 & .384 & .540 & .568 & .493 & .612 & .702 \\
\clineB{1-14}{2}
\end{tabular}
}
\caption{
Link prediction results for high-dimensional embeddings (best for $d\in\{200,400, 500\}$) in the filtered setting. 
{DistMult}, {ConvE} and {ComplEx} results are taken from \cite{dettmers2018convolutional}.
Best score in \textbf{bold} and best published \underline{underlined}. \textsc{AttE} and \textsc{AttH} have similar performance in the high-dimensional setting, performing competitively with or better than state-of-the-art methods on WN18RR, FB15k-237 and YAGO3-10.}
\label{tab:high_dim}
\end{table*}
\subsection{Results in high dimensions}\label{subsec:high_dim}
In high dimensions (Table \ref{tab:high_dim}), we compare against a variety of other models and achieve new SotA results on WN18RR and YAGO3-10, and third-best results on FB15k-237. 
As we expected, when the embedding dimension is large, Euclidean and hyperbolic embedding methods perform similarly across all datasets. 
We explain this behavior by noting that when the dimension is sufficiently large, both Euclidean and
hyperbolic spaces have enough capacity to represent complex hierarchies in KGs. 
This is further supported by Figure \ref{fig:learned_c}, which shows the learned absolute curvature versus the dimension. 
We observe that curvatures are close to zero in high dimensions, confirming our expectation that \textsc{RotH} with trainable curvatures learns a roughly Euclidean geometry in this setting. 

In contrast, fixed curvature degrades performance in high dimensions (Figure \ref{fig:sweep_d}), confirming the importance of trainable curvatures and its impact on precision and capacity (previously studied by \citep{sala2018representation}). 
Additionally, we show the embeddings' norms distribution in the Appendix (Figure \ref{fig:precision_plot}). 
Fixed curvature results in embeddings being clustered near the boundary of the ball while trainable curvatures adjusts the embedding space to better distribute points throughout the ball.
Precision issues that might arise with fixed curvature could also explain MurP's low performance in high dimensions. 
Trainable curvatures allow \textsc{RotH} to perform as well or better than previous methods in both low and high dimensions.

\subsection{Visualizations}
\begin{figure}[t]
\centering
\begin{subfigure}[b]{0.49\textwidth}
        \includegraphics[width=\textwidth]{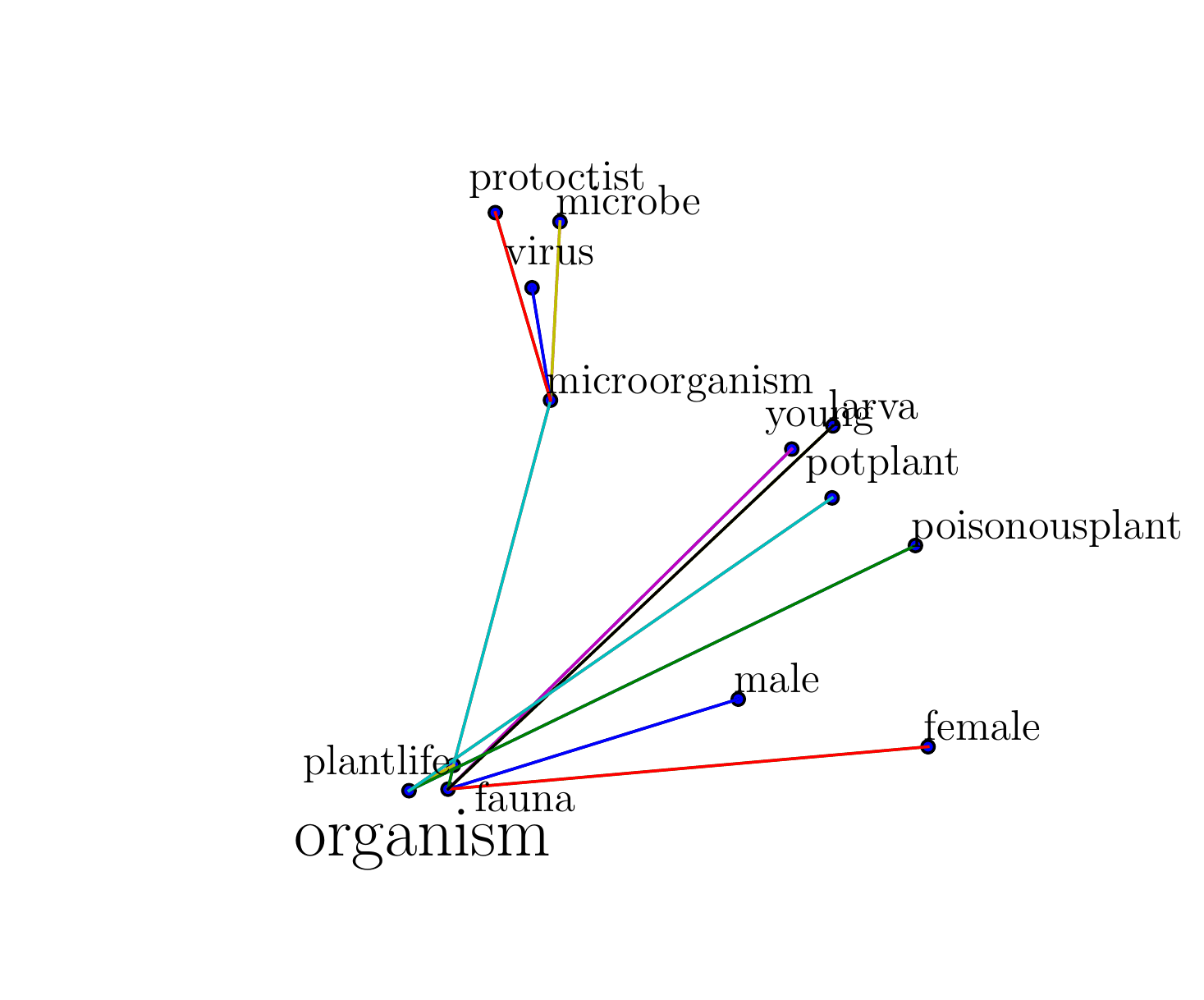}
        \caption{\textsc{RotE} embeddings.}
        \label{fig:vizrotE}
    \end{subfigure}
    \begin{subfigure}[b]{0.49\textwidth}
        \includegraphics[width=\textwidth]{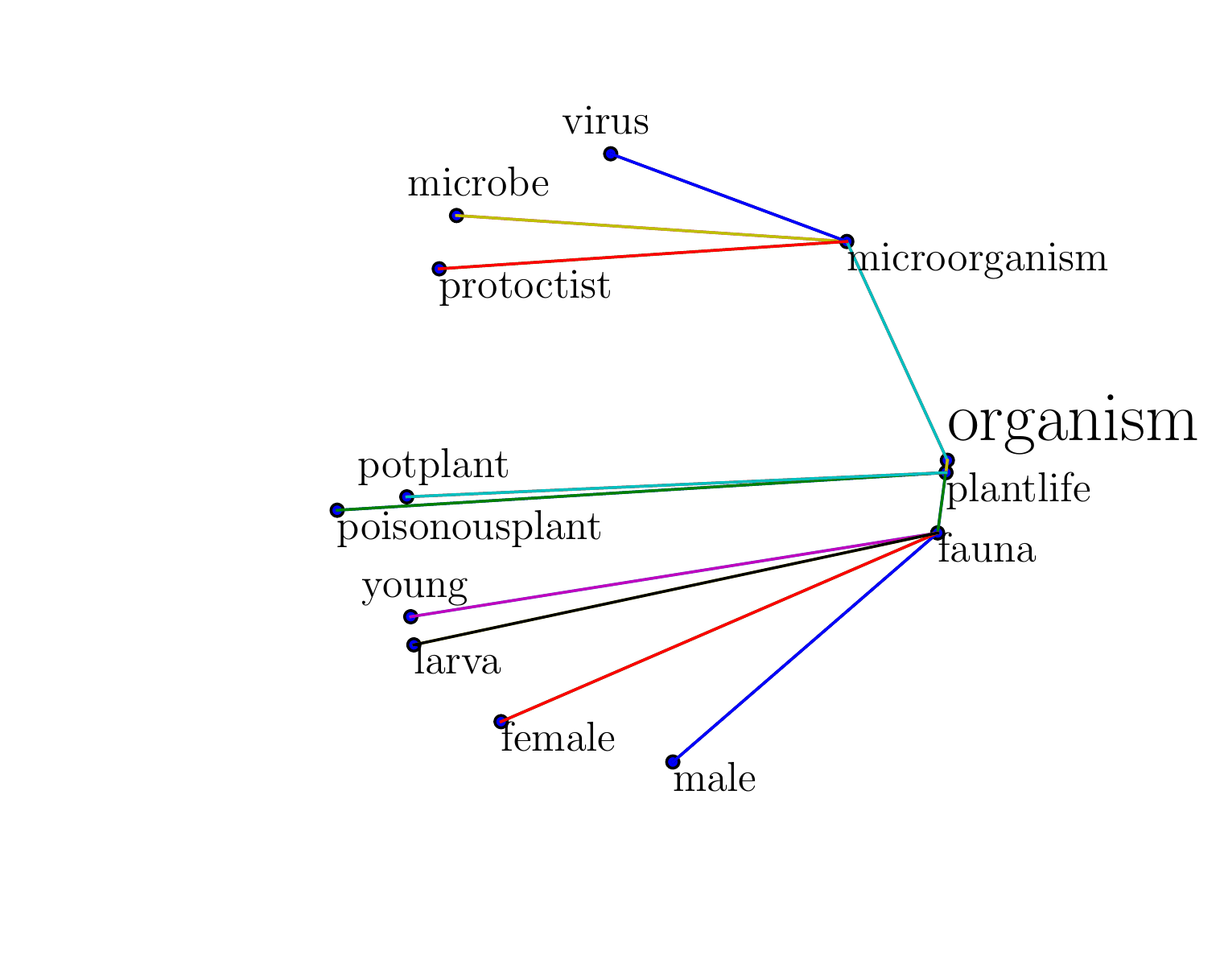}
        \caption{\textsc{RotH} embeddings.}
        \label{fig:vizrotH}
    \end{subfigure}
\caption{Visualizations of the embeddings learned by \textsc{RotE} and \textsc{RotH} on a sub-tree of WN18RR for the \textit{hypernym} relation.
In contrast to \textsc{RotE}, \textsc{RotH} preserves hierarchies by learning tree-like embeddings. 
}\label{fig:viz}
\end{figure}

In Figure \ref{fig:viz}, we visualize the embeddings learned by \textsc{RotE} versus \textsc{RotH} for a sub-tree of the \textit{organism} entity in WN18RR.
To better visualize the hierarchy, we apply $k$ inverse rotations for all nodes at level $k$ in the tree.

By contrast to \textsc{RotE}, \textsc{RotH} preserves the tree structure in the embedding space.
Furthermore, we note that \textsc{RotE} cannot simultaneously preserve the tree structure and make non-neighboring nodes far from each other. 
For instance, \textit{virus} should be far from \textit{male}, but preserving the tree structure (by going one level down in the tree) while making these two nodes far from each other is difficult in Euclidean space. 
In hyperbolic space, however, we observe that going one level down in the tree is achieved by translating embeddings towards the left. 
This pattern essentially illustrates the translation component in \textsc{RotH}, allowing the model to simultaneously preserve hierarchies while making non-neighbouring nodes far from each other. 

\section{Conclusion}
We introduce \model{}, a hyperbolic KG embedding model that leverages the expressiveness of hyperbolic space and attention-based geometric transformations to learn improved KG representations in low-dimensions.  
\model{} learns embeddings with trainable hyperbolic curvatures, allowing it to learn the right geometry for each relationship and generalize across multiple embedding dimensions. 
\model{} achieves new SotA on WN18RR and YAGO3-10, real-world KGs which exhibit hierarchical structures. 
Future directions for this work include exploring other tasks that might benefit from hyperbolic geometry, such as hypernym detection. 
The proposed attention-based transformations can also be extended to other geometric operations.

\section*{Acknowledgements}
We thank Avner May for their helpful feedback and discussions. We gratefully acknowledge the support of DARPA under Nos. FA86501827865 (SDH) and FA86501827882 (ASED); NIH under No. U54EB020405 (Mobilize), NSF under Nos. CCF1763315 (Beyond Sparsity), CCF1563078 (Volume to Velocity), and 1937301 (RTML); ONR under No. N000141712266 (Unifying Weak Supervision); the Moore Foundation, NXP, Xilinx, LETI-CEA, Intel, IBM, Microsoft, NEC, Toshiba, TSMC, ARM, Hitachi, BASF, Accenture, Ericsson, Qualcomm, Analog Devices, the Okawa Foundation, American Family Insurance, Google Cloud, Swiss Re, the HAI-AWS Cloud Credits for Research program, TOTAL, and members of the Stanford DAWN project: Teradata, Facebook, Google, Ant Financial, NEC, VMWare, and Infosys. The U.S. Government is authorized to reproduce and distribute reprints for Governmental purposes notwithstanding any copyright notation thereon. Any opinions, findings, and conclusions or recommendations expressed in this material are those of the authors and do not necessarily reflect the views, policies, or endorsements, either expressed or implied, of DARPA, NIH, ONR, or the U.S. Government.

\bibliography{references}

\begin{thebibliography}{34}
\expandafter\ifx\csname natexlab\endcsname\relax\def\natexlab#1{#1}\fi

\bibitem[{Bala{\v{z}}evi{\'c} et~al.(2019)Bala{\v{z}}evi{\'c}, Allen, and
  Hospedales}]{balavzevic2019multi}
Ivana Bala{\v{z}}evi{\'c}, Carl Allen, and Timothy Hospedales. 2019.
\newblock Multi-relational poincar\'e graph embeddings.
\newblock In \emph{Advances in Neural Information Processing Systems}, pages
  4465--4475.

\bibitem[{Balazevic et~al.(2019)Balazevic, Allen, and
  Hospedales}]{balazevic2019tucker}
Ivana Balazevic, Carl Allen, and Timothy Hospedales. 2019.
\newblock Tucker: Tensor factorization for knowledge graph completion.
\newblock In \emph{Proceedings of the 2019 Conference on Empirical Methods in
  Natural Language Processing and the 9th International Joint Conference on
  Natural Language Processing (EMNLP-IJCNLP)}, pages 5188--5197.

\bibitem[{Bansal et~al.(2019)Bansal, Juan, Ravi, and McCallum}]{bansal2019a2n}
Trapit Bansal, Da-Cheng Juan, Sujith Ravi, and Andrew McCallum. 2019.
\newblock A2n: Attending to neighbors for knowledge graph inference.
\newblock In \emph{Proceedings of the 57th Annual Meeting of the Association
  for Computational Linguistics}, pages 4387--4392.

\bibitem[{Bonnabel(2013)}]{bonnabel2013stochastic}
Silvere Bonnabel. 2013.
\newblock Stochastic gradient descent on {R}iemannian manifolds.
\newblock \emph{IEEE Transactions on Automatic Control}, 58(9):2217--2229.

\bibitem[{Bordes et~al.(2013)Bordes, Usunier, Garcia-Duran, Weston, and
  Yakhnenko}]{bordes2013translating}
Antoine Bordes, Nicolas Usunier, Alberto Garcia-Duran, Jason Weston, and Oksana
  Yakhnenko. 2013.
\newblock Translating embeddings for modeling multi-relational data.
\newblock In \emph{Advances in Neural Information Processing Systems}, pages
  2787--2795.

\bibitem[{Chami et~al.(2019)Chami, Ying, R{\'e}, and
  Leskovec}]{chami2019hyperbolic}
Ines Chami, Zhitao Ying, Christopher R{\'e}, and Jure Leskovec. 2019.
\newblock Hyperbolic graph convolutional neural networks.
\newblock In \emph{Advances in Neural Information Processing Systems}, pages
  4869--4880.

\bibitem[{Dettmers et~al.(2018)Dettmers, Minervini, Stenetorp, and
  Riedel}]{dettmers2018convolutional}
Tim Dettmers, Pasquale Minervini, Pontus Stenetorp, and Sebastian Riedel. 2018.
\newblock Convolutional {2D} knowledge graph embeddings.
\newblock In \emph{Thirty-Second AAAI Conference on Artificial Intelligence}.

\bibitem[{Duchi et~al.(2011)Duchi, Hazan, and Singer}]{duchi2011adaptive}
John Duchi, Elad Hazan, and Yoram Singer. 2011.
\newblock Adaptive subgradient methods for online learning and stochastic
  optimization.
\newblock \emph{Journal of Machine Learning Research}, 12(Jul):2121--2159.

\bibitem[{Ganea et~al.(2018)Ganea, B{\'e}cigneul, and
  Hofmann}]{ganea2018hyperbolicNN}
Octavian Ganea, Gary B{\'e}cigneul, and Thomas Hofmann. 2018.
\newblock Hyperbolic neural networks.
\newblock In \emph{Advances in Neural Information Processing Systems}.

\bibitem[{Gu et~al.(2019)Gu, Sala, Gunel, and R{\'e}}]{gu2019mixed}
Albert Gu, Fred Sala, Beliz Gunel, and Christopher R{\'e}. 2019.
\newblock Learning mixed-curvature representations in product spaces.
\newblock In \emph{International Conference on Learning Representations}.

\bibitem[{Ji et~al.(2015)Ji, He, Xu, Liu, and Zhao}]{ji2015knowledge}
Guoliang Ji, Shizhu He, Liheng Xu, Kang Liu, and Jun Zhao. 2015.
\newblock Knowledge graph embedding via dynamic mapping matrix.
\newblock In \emph{Proceedings of the 53rd Annual Meeting of the Association
  for Computational Linguistics and the 7th International Joint Conference on
  Natural Language Processing (Volume 1: Long Papers)}, pages 687--696.

\bibitem[{Kingma and Ba(2015)}]{kingma2014adam}
Diederik~P Kingma and Jimmy Ba. 2015.
\newblock Adam: A method for stochastic optimization.
\newblock In \emph{International Conference for Learning Representations}.

\bibitem[{Krackhardt(1994)}]{krackhardt2014graph}
David Krackhardt. 1994.
\newblock Graph theoretical dimensions of informal organizations.
\newblock In \emph{Computational organization theory}, pages 107--130.
  Psychology Press.

\bibitem[{Lacroix et~al.(2018)Lacroix, Usunier, and
  Obozinski}]{lacroix2018canonical}
Timoth{\'e}e Lacroix, Nicolas Usunier, and Guillaume Obozinski. 2018.
\newblock Canonical tensor decomposition for knowledge base completion.
\newblock \emph{International Conference on Machine Learning}.

\bibitem[{Lin et~al.(2015)Lin, Liu, Sun, Liu, and Zhu}]{lin2015learning}
Yankai Lin, Zhiyuan Liu, Maosong Sun, Yang Liu, and Xuan Zhu. 2015.
\newblock Learning entity and relation embeddings for knowledge graph
  completion.
\newblock In \emph{Twenty-ninth AAAI Conference on Artificial Intelligence}.

\bibitem[{Liu et~al.(2019)Liu, Nickel, and Kiela}]{liu2019hyperbolic}
Qi~Liu, Maximilian Nickel, and Douwe Kiela. 2019.
\newblock Hyperbolic graph neural networks.
\newblock In \emph{Advances in Neural Information Processing Systems}, pages
  8228--8239.

\bibitem[{Mahdisoltani et~al.(2013)Mahdisoltani, Biega, and
  Suchanek}]{mahdisoltani2013yago3}
Farzaneh Mahdisoltani, Joanna Biega, and Fabian~M Suchanek. 2013.
\newblock Yago3: A knowledge base from multilingual wikipedias.

\bibitem[{Miller(1995)}]{miller1995wordnet}
George~A Miller. 1995.
\newblock Wordnet: a lexical database for english.
\newblock \emph{Communications of the ACM}, 38(11):39--41.

\bibitem[{Nathani et~al.(2019)Nathani, Chauhan, Sharma, and Kaul}]{KBGAT2019}
Deepak Nathani, Jatin Chauhan, Charu Sharma, and Manohar Kaul. 2019.
\newblock Learning attention-based embeddings for relation prediction in
  knowledge graphs.
\newblock In \emph{Proceedings of the 57th Annual Meeting of the Association
  for Computational Linguistics}. Association for Computational Linguistics.

\bibitem[{Nguyen et~al.(2018)Nguyen, Nguyen, Nguyen, and Phung}]{Nguyen2018}
Dai~Quoc Nguyen, Tu~Dinh Nguyen, Dat~Quoc Nguyen, and Dinh Phung. 2018.
\newblock {A Novel Embedding Model for Knowledge Base Completion Based on
  Convolutional Neural Network}.
\newblock In \emph{Proceedings of the 16th Annual Conference of the North
  American Chapter of the Association for Computational Linguistics: Human
  Language Technologies (NAACL-HLT)}, pages 327--333.

\bibitem[{Nickel et~al.(2011)Nickel, Tresp, and Kriegel}]{nickel2011three}
Maximilian Nickel, Volker Tresp, and Hans-Peter Kriegel. 2011.
\newblock A three-way model for collective learning on multi-relational data.
\newblock In \emph{International Conference on Machine Learning}, pages
  809--816. Omnipress.

\bibitem[{Nickel and Kiela(2017)}]{Nickel2017-mw}
Maximillian Nickel and Douwe Kiela. 2017.
\newblock Poincar{\'e} embeddings for learning hierarchical representations.
\newblock In \emph{Advances in Neural Information Processing Systems}, pages
  6338--6347.

\bibitem[{Robbin and Salamon()}]{robbin2011introduction}
Joel~W Robbin and Dietmar~A Salamon.
\newblock Introduction to differential geometry.

\bibitem[{Sala et~al.(2018)Sala, De~Sa, Gu, and R\'e}]{sala2018representation}
Frederic Sala, Chris De~Sa, Albert Gu, and Christopher R\'e. 2018.
\newblock Representation tradeoffs for hyperbolic embeddings.
\newblock In \emph{International Conference on Machine Learning}, pages
  4457--4466.

\bibitem[{Schlichtkrull et~al.(2018)Schlichtkrull, Kipf, Bloem, Van Den~Berg,
  Titov, and Welling}]{schlichtkrull2018modeling}
Michael Schlichtkrull, Thomas~N Kipf, Peter Bloem, Rianne Van Den~Berg, Ivan
  Titov, and Max Welling. 2018.
\newblock Modeling relational data with graph convolutional networks.
\newblock In \emph{European Semantic Web Conference}, pages 593--607. Springer.

\bibitem[{Suchanek et~al.(2007)Suchanek, Kasneci, and
  Weikum}]{suchanek2007yago}
Fabian~M Suchanek, Gjergji Kasneci, and Gerhard Weikum. 2007.
\newblock Yago: a core of semantic knowledge.
\newblock In \emph{Proceedings of the 16th international conference on World
  Wide Web}, pages 697--706. ACM.

\bibitem[{Sun et~al.(2019)Sun, Deng, Nie, and Tang}]{sun2019rotate}
Zhiqing Sun, Zhi-Hong Deng, Jian-Yun Nie, and Jian Tang. 2019.
\newblock Rotate: Knowledge graph embedding by relational rotation in complex
  space.
\newblock In \emph{International Conference on Learning Representations}.

\bibitem[{Tifrea et~al.(2019)Tifrea, B{\'e}cigneul, and
  Ganea}]{tifrea2018poincar}
Alexandru Tifrea, Gary B{\'e}cigneul, and Octavian-Eugen Ganea. 2019.
\newblock Poincar\'e {GloVe}: Hyperbolic word embeddings.
\newblock In \emph{International Conference on Learning Representations}.

\bibitem[{Toutanova and Chen(2015)}]{toutanova2015observed}
Kristina Toutanova and Danqi Chen. 2015.
\newblock Observed versus latent features for knowledge base and text
  inference.
\newblock In \emph{Proceedings of the 3rd Workshop on Continuous Vector Space
  Models and their Compositionality}, pages 57--66.

\bibitem[{Trouillon et~al.(2016)Trouillon, Welbl, Riedel, Gaussier, and
  Bouchard}]{trouillon2016complex}
Th{\'e}o Trouillon, Johannes Welbl, Sebastian Riedel, {\'E}ric Gaussier, and
  Guillaume Bouchard. 2016.
\newblock Complex embeddings for simple link prediction.
\newblock In \emph{International Conference on Machine Learning}, pages
  2071--2080.

\bibitem[{Wang et~al.(2014)Wang, Zhang, Feng, and Chen}]{wang2014knowledge}
Zhen Wang, Jianwen Zhang, Jianlin Feng, and Zheng Chen. 2014.
\newblock Knowledge graph embedding by translating on hyperplanes.
\newblock In \emph{Twenty-Eighth AAAI Conference on Artificial Intelligence}.

\bibitem[{Xu and Li(2019)}]{xu2019relation}
Canran Xu and Ruijiang Li. 2019.
\newblock Relation embedding with dihedral group in knowledge graph.
\newblock In \emph{Proceedings of the 57th Annual Meeting of the Association
  for Computational Linguistics}. Association for Computational Linguistics.

\bibitem[{Yang et~al.(2015)Yang, Yih, He, Gao, and Deng}]{yang2014embedding}
Bishan Yang, Wen-tau Yih, Xiaodong He, Jianfeng Gao, and Li~Deng. 2015.
\newblock Embedding entities and relations for learning and inference in
  knowledge bases.
\newblock In \emph{International Conference on Learning Representations}.

\bibitem[{Zhang et~al.(2019)Zhang, Tay, Yao, and Liu}]{zhang2019quaternion}
Shuai Zhang, Yi~Tay, Lina Yao, and Qi~Liu. 2019.
\newblock Quaternion knowledge graph embeddings.
\newblock In \emph{Advances in Neural Information Processing Systems}, pages
  2731--2741.

\end{thebibliography}
\bibliographystyle{acl2020/acl_natbib}

\clearpage
\appendix
\section{Appendix}
Below, we provide additional details. 
We start by providing the formula for the hyperbolic analogue of addition that we use, along with additional hyperbolic geometry background. 
Next, we provide more information about the metrics that are used to determine how hierarchical a dataset is.
Afterwards, we give additional experimental details, including the table of hyperparameters and further details on tangent space optimization. 
Lastly, we include an additional comparison against the Dihedral model \cite{xu2019relation}.

\subsection{M\"obius addition}\label{sec:appendix_hyp}
The M\"obius addition operation \cite{ganea2018hyperbolicNN} has the closed-form expression:
\begin{align*}
    \mathbf{x}\oplus^c\mathbf{y}&=\frac{\alpha_{\mathbf{x}\mathbf{y}}\mathbf{x}+\beta_{\mathbf{x}\mathbf{y}}\mathbf{y}}{1 + 2c\mathbf{x}^T\mathbf{y} + c^2||\mathbf{x}||^2||\mathbf{y}||^2},\\
    \text{where}\ \ \alpha_{\mathbf{x}\mathbf{y}}&=1+2c \mathbf{x}^T\mathbf{y} + c||\mathbf{y}||^2,\\
    \text{and}\ \ \beta_{\mathbf{x}\mathbf{y}}&=1-c||\mathbf{x}||^2.
\end{align*}
In contrast to Euclidean addition, it is neither commutative nor associative. 
However, it provides an analogue through the lens of parallel transport: 
given two points $\mathbf{x}, \mathbf{y}$ and a vector $\mathbf{v}$ in $\mathcal{T}^c_\mathbf{x}$, there is a unique vector in $\mathcal{T}^c_\mathbf{y}$ which creates the same angle as $\mathbf{v}$ with the direction of the geodesic (shortest path) connecting $\mathbf{x}$ to $\mathbf{y}$. 
This map is the parallel transport $P^c_{\mathbf{x}\rightarrow \mathbf{y}}(\cdot)$; Euclidean parallel transport is the standard Euclidean addition. 
Analogously, the M\"obius addition satisfies \cite{ganea2018hyperbolicNN}:
$\mathbf{x}\oplus^c\mathbf{y}=\mathrm{exp}_\mathbf{x}^c(P^c_{\mathbf{0}\rightarrow \mathbf{x}}(\mathrm{log}_\mathbf{\mathbf{0}}^c(\mathbf{y})))$.

\subsection{Hierarchy estimates}\label{subsec:curvature_est}
We use two metrics to estimate how hierarchical a relation is: the curvature estimate $\xi_G$ and the Krackhardt hierarchy score $\text{Khs}_G$. While the curvature estimate captures global hierarchical behaviours (how much the graph is tree-like when zooming-out), the Krackhardt score captures a more local behaviour (how many small loops the graph has). See Figure \ref{fig:hier_measures} for examples. 

\paragraph{Curvature estimate} To estimate the curvature of a relation $r$, we restrict to the undirected graph $G_r$ spanned by the edges labeled as $r$. Following \cite{gu2019mixed}, let $\xi_{G_r}(a,b,c)$ be the curvature estimate of a triangle in $G_r$ with vertices $\{a,b,c\}$, which is given by:
\begin{flalign*}
\xi_{G_r}(a,b,c) &=\frac{1}{2d_{G_r}(a,m)} \big( d_{G_r}(a,m)^2
\\&
+ d_{G_r}(b,c)^2/4 
\\&
- (d_{G_r}(a,b)^2 +d_{G_r}(a,c)^2)/2 \big),
\end{flalign*}
where $m$ is the midpoint of the shortest path connecting $b$ to $c$. This estimate is positive for triangles in circles, negative for triangles in trees, and zero for triangles in lines. Moreover, for a triangle in a Riemannian manifold $M$, $\xi_M(a,b,c)$ estimates the sectional curvature of the plane on which the triangle lies (see \cite{gu2019mixed} for more details). Let $m_r$ be the total number of connected components in $G_r$. We sample $1000\ w_{i,r}$ triangles from each connected component $c_{i,r}$ of $G_r$ where $w_{i,r}= \frac{N_{i,r}^3}{\sum_{i=1}^{m_r} N_{i,r}^3}$, and $N_{i,r}$ is the number of nodes in the component $c_{i,r}$. $\xi_{G_r}$ is the mean of the estimated curvatures of the sampled triangles. For the full graph, we take the weighted average of the relation curvatures $\xi_{G_r}$ with respect to the weights $\frac{\sum_{i=1}^{m_r} N_{i,r}^3}{\sum_r\sum_{i=1}^{m_r} N_{i,r}^3}.$

\paragraph{Krackhardt hierarchy score}
For the directed graph $G_r$ spanned by the relation $r$, we let $R$ be the adjacency matrix ($R_{i,j}=1$ if there is an edge from node $i$ to node $j$ and $0$ otherwise). 
Then: 
$$\text{Khs}_{G_r} = \frac{\sum_{i,j=1}^n R_{i,j}(1-R_{j,i})}{\sum_{i,j=1}^n R_{i,j}}.$$
See \cite{krackhardt2014graph} for more details. 
We note that for fully observed symmetric relations (each edge is in a two-edge loop), $\text{Khs}_{G_r} = 0$ while for anti-symmetric relations (no small loops), $\text{Khs}_{G_r} = 1$.

 \begin{figure}[t]
\centering
\begin{subfigure}[b]{\textwidth}
    \includegraphics[width=\textwidth]{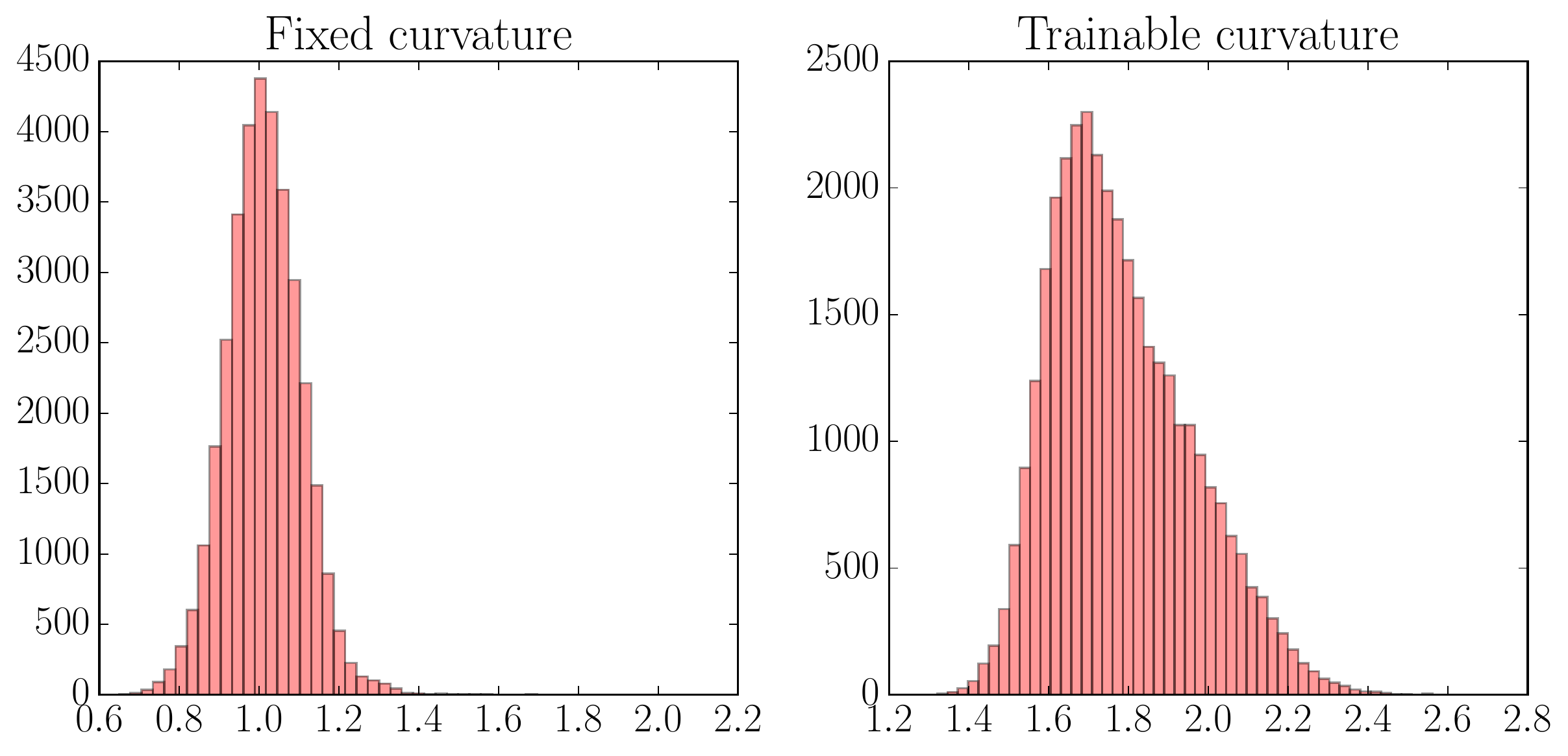}
    \end{subfigure}
\caption{Histogram of embeddings norm learned with fixed and trainable curvatures for the hypernym relation in WN18RR.
}\label{fig:precision_plot}
\end{figure} 
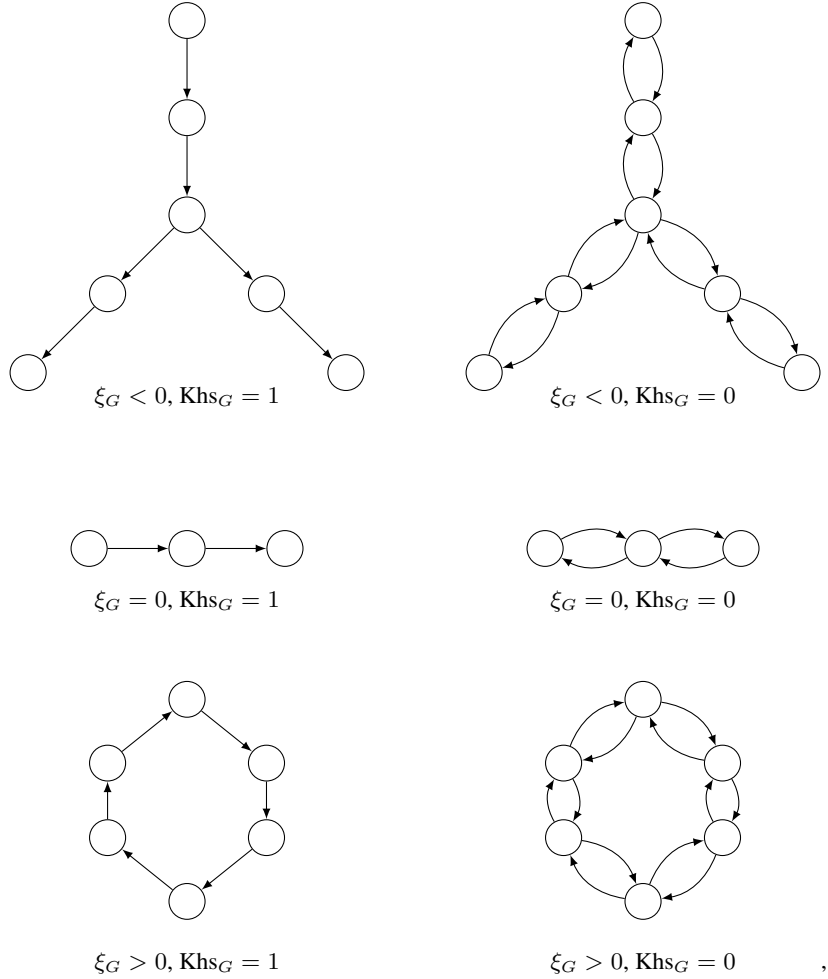
\begin{figure*}[th]
\centering
\small
\tikzset{main node/.style={shape=circle,draw,minimum size=1.5em}}
\tikzset{edge/.style = {->,> = latex}}
  \begin{tikzpicture}
    \node[main node] (1) {};
    \node[main node] (2) [below = 0.8cm of 1]   {};
    \node[main node] (3) [below = 0.8cm of 2]  {};
    \node[main node] (4) [below left = 0.7cm and 0.7cm of 3]   {};
    \node[main node] (5) [below left = 0.7cm and 0.7cm of 4]  {} ;
    \node[main node] (6) [below right = 0.7cm and 0.7cm of 3]  {} ;
    \node[main node] (7) [below right = 0.7cm and 0.7cm of 6]  {} ;
    \node[anchor=west] [below = 4.5cm of 1] {$\xi_G <0$, $\text{Khs}_G = 1$};

    \draw[edge](1) to (2);
    \draw[edge](2) to (3);
    \draw[edge](3) to (4);
    \draw[edge](4) to (5);
    \draw[edge](3) to (6);
    \draw[edge](6) to (7);
    \begin{scope}[xshift=6cm]
    \node[main node] (1) {};
    \node[main node] (2) [below = 0.8cm of 1]   {};
    \node[main node] (3) [below = 0.8cm of 2]  {};
    \node[main node] (4) [below left = 0.7cm and 0.7cm of 3]   {};
    \node[main node] (5) [below left = 0.7cm and 0.7cm of 4]  {} ;
    \node[main node] (6) [below right = 0.7cm and 0.7cm of 3]  {} ;
    \node[main node] (7) [below right = 0.7cm and 0.7cm of 6]  {} ;
    
    \draw[edge](1) to[bend left] (2);
    \draw[edge](2) to[bend left] (3);
    \draw[edge](3) to[bend left] (4);
    \draw[edge](4) to[bend left] (5);
    \draw[edge](3) to[bend left] (6);
    \draw[edge](6) to[bend left] (7);
    
    \draw[edge](2) to[bend left] (1);
    \draw[edge](3) to[bend left] (2);
    \draw[edge](4) to[bend left] (3);
    \draw[edge](5) to[bend left] (4);
    \draw[edge](6) to[bend left] (3);
    \draw[edge](7) to[bend left] (6);
    
    \node[anchor=west] [below = 4.5cm of 1] {$\xi_G <0$, $\text{Khs}_G = 0$};
    \end{scope}
    
    \begin{scope}[yshift=-7cm]
    \node[main node] (1) {};
    \node[main node] (2) [left = 0.8cm of 1]   {};
    \node[main node] (3) [right = 0.8cm of 1]  {};
    
    \draw[edge](2) to (1);
    \draw[edge](1) to (3);
    
    \node[anchor=west] [below = 0.2cm of 1] {$\xi_G =0$, $\text{Khs}_G = 1$};
    \end{scope}
    \begin{scope}[yshift=-7cm, xshift = 6cm]
    \node[main node] (1) {};
    \node[main node] (2) [left = 0.8cm of 1]   {};
    \node[main node] (3) [right = 0.8cm of 1]  {};
    
    \draw[edge](2) to[bend left] (1);
    \draw[edge](1) to[bend left] (3);
    
    \draw[edge](1) to[bend left] (2);
    \draw[edge](3) to[bend left] (1);

    \node[anchor=west] [below = 0.2cm of 1] {$\xi_G =0$, $\text{Khs}_G = 0$};
    \end{scope}
    
    \begin{scope}[yshift=-9cm]
    \node[main node] (1) {};
    \node[main node] (2) [below left = 0.5cm and 0.7cm of 1]   {};
    \node[main node] (3) [below right = 0.5cm and 0.7cm of 1]  {};
    \node[main node] (4) [below = 0.5cm of 2]  {};
    \node[main node] (5) [below = 0.5cm of 3]  {};
    \node[main node] (6) [below right = 0.5cm and 0.7cm of 4]  {};
    
    \draw[edge](2) to (1);
    \draw[edge](1) to (3);
    \draw[edge](4) to (2);
    \draw[edge](3) to (5);
    \draw[edge](6) to (4);
    \draw[edge](5) to (6);
    
    \node[anchor=west] [below = 3cm of 1] {$\xi_G >0$, $\text{Khs}_G = 1$};
    \end{scope}
    
    \begin{scope}[yshift=-9cm, xshift=6cm]
    \node[main node] (1) {};
    \node[main node] (2) [below left = 0.5cm and 0.7cm of 1]   {};
    \node[main node] (3) [below right = 0.5cm and 0.7cm of 1]  {};
    \node[main node] (4) [below = 0.5cm of 2]  {};
    \node[main node] (5) [below = 0.5cm of 3]  {};
    \node[main node] (6) [below right = 0.5cm and 0.7cm of 4]  {};
    
    \draw[edge](2) to[bend left] (1);
    \draw[edge](1) to[bend left] (3);
    \draw[edge](4) to[bend left] (2);
    \draw[edge](3) to[bend left] (5);
    \draw[edge](6) to[bend left] (4);
    \draw[edge](5) to[bend left] (6);

    \draw[edge](1) to[bend left] (2);
    \draw[edge](3) to[bend left] (1);
    \draw[edge](2) to[bend left] (4);
    \draw[edge](5) to[bend left] (3);
    \draw[edge](4) to[bend left] (6);
    \draw[edge](6) to[bend left] (5);
    
    \node[anchor=west] [below = 3cm of 1] {$\xi_G >0$, $\text{Khs}_G = 0$};
    \end{scope}

\end{tikzpicture}'
\caption{The curvature estimate $\xi_G$ and the Krackhardt hierarchy score $\text{Khs}_G$ for several simple graphs. The top-left graph is the most hierarchical, while the bottom-right graph is the least hierarchical.} \label{fig:hier_measures}
\end{figure*}    
\begin{table}[t]
\centering
\resizebox{\textwidth}{!}{\renewcommand{\arraystretch}{1.1}
\begin{tabular}{lcccccc}
\clineB{1-7}{2}
 & \multicolumn{2}{c}{WN18RR} & \multicolumn{2}{c}{FB15k-237} & \multicolumn{2}{c}{YAGO3-10} \\
{Model} & MRR & H@10 & MRR & H@10 & MRR & H@10 \\
\hline
Dihedral & .486 & 557 & .300 & .496 & .388 & .573 \\
\textsc{AttE} & \textbf{.490} & \textbf{.581} & \textbf{.351} & \textbf{.543} & \textbf{.575} & \textbf{.709} \\
\clineB{1-7}{2}
\end{tabular}
\caption{Comparison of Dihedral and \textsc{AttE} in high-dimensions.} 
\label{tab:dihedral}
}
\end{table}
\begin{table*}[t]
\centering
\small
\begin{tabular}{l|clcccc}
\clineB{1-7}{2}
{Dataset} &  embedding dimension & model & {learning rate} & {optimizer} & {batch size} & {negative samples} \\
\clineB{1-7}{2}

\multirow{12}{*}{{WN18RR}} & \multirow{6}{*}{{32}} & \textsc{RefE} & 0.001 & Adam & 100 & 250 \\
&  & \textsc{RotE} & 0.001 & Adam & 100 & 250 \\
&  & \textsc{AttE}& 0.001 & Adam & 100 & 250\\
\cline{3-7} 
&  & \textsc{RefH} & 0.0005 & Adam & 250 & 250 \\
&  & \textsc{RotH} & 0.0005 & Adam & 500 & 50 \\
&  & \textsc{AttH} & 0.0005 & Adam & 500 & 50 \\
\cline{2-7}
& \multirow{6}{*}{{500}} & \textsc{RefE} & 0.1 & Adagrad & 500 & 50 \\
&  & \textsc{RotE} & 0.001 & Adam & 100 & 500 \\
&  & \textsc{AttE} & 0.001 & Adam & 1000 & 50 \\
\cline{3-7} 
&  & \textsc{RefH} & 0.05 & Adagrad & 500 & 50 \\
&  & \textsc{RotH} & 0.001 & Adam & 1000 & 50 \\
&  & \textsc{AttH} & 0.001 & Adam & 1000 & 50 \\
\clineB{1-7}{2}
\multirow{12}{*}{{FB15k-237}} & \multirow{6}{*}{{32}} & \textsc{RefE} & 0.075 & Adagrad & 250 & 250 \\
&  & \textsc{RotE} & 0.05 & Adagrad & 500 & 50 \\
&  & \textsc{AttE} & 0.05 & Adagrad & 500 & 50 \\
\cline{3-7} 
&  & \textsc{RefH} & 0.05 & Adagrad & 500 & 250 \\
&  & \textsc{RotH} & 0.1 & Adagrad & 100 & 50 \\
&  & \textsc{AttH} & 0.05 & Adagrad & 500 & 100 \\
\cline{2-7}
& \multirow{6}{*}{{500}} & \textsc{RefE} & 0.05 & Adagrad & 500 & 50 \\
&  & \textsc{RotE} & 0.05 & Adagrad & 100 & 50\\
&  & \textsc{AttE} & 0.05 & Adagrad & 500 & 50 \\
\cline{3-7} 
&  & \textsc{RefH} & 0.05 & Adagrad & 500 & 50\\
&  & \textsc{RotH} & 0.05 & Adagrad & 1000 & 50 \\
&  & \textsc{AttH} & 0.05 & Adagrad & 500 & 50 \\
\clineB{1-7}{2}
\multirow{12}{*}{{YAGO3-10}} & \multirow{6}{*}{{32}} & \textsc{RefE} & 0.005 & Adam & 2000 & NA \\
&  & \textsc{RotE} & 0.005 & Adam & 2000 & NA \\
&  & \textsc{AttE} & 0.005 & Adam & 2000 & NA\\
\cline{3-7} 
&  & \textsc{RefH} & 0.005 & Adam & 1000 & NA \\
&  & \textsc{RotH} & 0.001 & Adam & 1000 & NA\\
&  & \textsc{AttH} & 0.001 & Adam & 1000 & NA \\
\cline{2-7}
& \multirow{6}{*}{{500}} & \textsc{RefE} & 0.005 & Adam & 4000 & NA \\
&  & \textsc{RotE} & 0.005 & Adam & 4000 & NA\\
&  & \textsc{AttE} & 0.005 & Adam & 2000 & NA \\
\cline{3-7} 
&  & \textsc{RefH} & 0.001 & Adam & 1000 & NA \\
&  & \textsc{RotH} & 0.0005 & Adam & 1000 & NA \\
&  & \textsc{AttH} & 0.0005 & Adam & 1000 & NA \\
\clineB{1-7}{2}
\end{tabular}
\caption{Best hyperparameters in low- and high-dimensional settings. NA negative samples indicates that the full cross-entropy loss is used, without negative sampling.} 
\label{tab:hyper_param}
\end{table*}

\subsection{Experimental details}\label{subsec:appendix_param}
For all our Euclidean and hyperbolic models, we conduct a hyperparameter search for the learning rate, optimizer (Adam \cite{kingma2014adam} or Adagrad \cite{duchi2011adaptive}), negative sample size and batch size. 
We train each model for 500 epochs and use early stopping after 100 epochs if the validation MRR stops increasing. 
We report the best hyperparameters for each dataset in Table \ref{tab:hyper_param}. 

\subsection{Tangent space optimization}\label{subsec:tangent_optim}
Optimization in hyperbolic space normally requires Riemannian Stochastic Gradient Descent (RSGD) \cite{bonnabel2013stochastic}, as was used in MuRP. 
RSGD is challenging in practice. 
Instead, we use tangent space optimization \cite{chami2019hyperbolic}.
We define all the \model{} parameters in the tangent space at the origin (our parameter space), optimize embeddings using standard Euclidean techniques, and use the exponential map to recover the hyperbolic parameters.

Note that tangent space optimization is an exact procedure, which does not incur losses in representational power. 
This is the case in hyperbolic space specifically because of a completeness property: there is always a global bijection between the tangent space and the manifold.  

Concretely, \model{} optimizes the entity and relationship embeddings $(\mathbf{e}^E_v)_{v\in\mathcal{V}}$ and $(\mathbf{r}^E_r)_{r\in\mathcal{R}}$, which are mapped to the Poincar\'e ball with:
\begin{equation}
    \mathbf{e}_v^H=\mathrm{exp}_\mathbf{0}^{c_r}(\mathbf{e}_v^E)  \ \ \text{and}\ \
    \mathbf{r}_r^H=\mathrm{exp}_\mathbf{0}^{c_r}(\mathbf{r}_r^E),\label{eq:mapping}
\end{equation}
The trainable model parameters are then $\{(\Theta_r, \Phi_r, \mathbf{r}_r^E, \mathbf{a}_r, c_r)_{r\in\mathcal{R}},(\mathbf{e}_v^E, b_v)_{v\in\mathcal{V}}\}$, which are all Euclidean parameters that can be learned using standard Euclidean optimization techniques. 
\subsection{Comparison to Dihedral}
We compare the performance of Dihedral \cite{xu2019relation} versus that of \textsc{AttE} in Table \ref{tab:dihedral}. 
Both methods combine rotations and reflections, but our approach learns attention-based transformations, while Dihedral learns a single parameter to determine which transformation to use. 
\textsc{AttE} significantly outperforms Dihedral on all datasets, suggesting that using attention-based representations is important in order to learn the right geometric transformation for each relation.

\end{document}